\documentclass[lettersize,journal]{IEEEtran}
\usepackage{amsmath,amsfonts}
\usepackage{algorithmic}
\usepackage{algorithm}
\usepackage{array}
\usepackage[caption=false,font=footnotesize,labelfont=sf,textfont=sf]{subfig}
\usepackage{textcomp}
\usepackage{stfloats}
\usepackage{url}
\usepackage{verbatim}
\usepackage{graphicx}
\usepackage{cite}
\usepackage[table,xcdraw]{xcolor}
\usepackage{tabularx}
\usepackage{colortbl}
\usepackage{adjustbox}
\usepackage{multirow} 
\usepackage[citecolor=blue, colorlinks]{hyperref}

\usepackage{soul}
\newcommand{\ADD}[1]{\textcolor{black}{\textbf{}#1}}
\newcommand{\CZADD}[1]{\textcolor{black}{\textbf{}#1}}

\begin{document}

\title{RBFIM: Perceptual Quality Assessment for Compressed Point Clouds Using Radial Basis Function Interpolation}

\author{Zhang Chen, Shuai Wan,~\IEEEmembership{Member,~IEEE,} Siyu Ren, Fuzheng Yang,~\IEEEmembership{Member,~IEEE}, Mengting Yu, \\ and Junhui Hou,~\IEEEmembership{Senior Member,~IEEE}
\thanks{Zhang Chen and Mengting Yu are with the School of Electronics and Information, Northwestern Polytechnical University, Xi’an 710129, China (e-mail: chenzhang@mail.nwpu.edu.cn;yumengting@mail.nwpu.edu.cn).}
\thanks{Shuai Wan is with the School of Electronics and Information, Northwestern Polytechnical University, Xi’an 710129, China, and also with the School of Engineering, Royal Melbourne Institute of Technology, Melbourne, VIC 3001, Australia (e-mail: swan@nwpu.edu.cn).}
\thanks{Siyu Ren and Junhui Hou are with the Department of Computer Science, City University of Hong Kong, Hong Kong SAR (e-mail: siyuren2-c@my.cityu.edu.hk; jh.hou@cityu.edu.hk).}
\thanks{Fuzheng Yang is with the School of Telecommunication Engineering, Xidian University, Xi’an 710071, China (e-mail: fzhyang@mail.xidian.edu.cn).}
\thanks{This work was supported in part by the TCL Science and Technology Innovation Fund, in part by the NSFC Fund 62371358, in part by the NSFC Excellent Young Scientists Fund 62422118 and in part by the Hong Kong Research Grants Council under Grants 11219422 and 11219324.}
}

\markboth{}%
{Shell \MakeLowercase{ }
}
\maketitle
\begin{abstract}
One of the main challenges in point cloud compression (PCC) is how to evaluate the perceived distortion so that the codec can be optimized for perceptual quality. Current standard practices in PCC highlight a primary issue: while single-feature metrics are widely used to assess compression distortion, the classic method of searching point-to-point nearest neighbors frequently fails to adequately \ADD{build} precise correspondences between point clouds, resulting in an ineffective capture of human perceptual features. To overcome \ADD{the related} limitations, we propose a novel assessment method called RBFIM, utilizing radial basis function (RBF) interpolation to convert discrete point features into a continuous feature function for the distorted point cloud. By substituting the geometry coordinates of the original point cloud into the feature function, we obtain the bijective sets of point features. This \ADD{enables an establishment of} precise corresponding features between distorted and original point clouds and significantly improves the accuracy of quality assessments. Moreover, this method \ADD{avoids} the complexity caused by bidirectional searches. Extensive experiments on multiple \ADD{subjective quality datasets of compressed point clouds} demonstrate that our RBFIM excels in addressing human perception tasks, thereby providing robust support for PCC optimization efforts.
\end{abstract}
\section{Introduction}
\IEEEPARstart{T}{he} booming significance of three-dimensional point cloud data in various cutting-edge applications, including augmented reality\cite{ref1}, autonomous navigation\cite{ref2}, and digital heritage conservation\cite{ref3}, underscores the necessity of effective data management techniques. Notably, point clouds encapsulate voluminous spatial and depth information, making their storage and transmission particularly resource-intensive\cite{ref4}. Consequently, related compression methods are indispensable\cite{ref5}\ADD{,}\cite{ref6}, yet they must be complemented by equally effective mechanisms for evaluating the perceptual quality of compressed outputs to ensure practical usability and user satisfaction\ADD{\cite{ref7}\cite{ref8}\cite{reviewer 2-2}\cite{reviewer 2-4}\cite{reviewer 2-5}.}\par
Research in point cloud quality assessment (PCQA) highlights two main directions: evaluation based on two-dimensional projections\cite{ref9}\cite{ref10}\cite{ref11} and quality assessment of three-dimensional spatial features\cite{ref12}\cite{ref13}\cite{ref14}. Algorithms using two-dimensional projection project point clouds onto two-dimensional planes, where well-established image quality assessment methods are reused or enhanced to evaluate the quality of the projected images from various viewpoints \ADD{\cite{reviewer1-1, reviewer1-2}}. Then\ADD{,} the quality of point clouds is achieved by appropriately pooling the quality of projected images. Such approaches accomplish the task of assessing the quality of three-dimensional point clouds by leveraging the principles of image quality assessment\cite{ref15}. However, algorithms resorting to two-dimensional projection face the challenges of occlusion and voids generated by projection, which have a significant impact on the evaluation results\cite{ref16}. Moreover, the projection of the entire point cloud from three-dimensional to two-dimensional is costly.\par
On the other hand, quality evaluation can be performed by using three-dimensional spatial features instead of projection. For example, in the current point cloud compression (PCC) standards\ADD{,} point-to-point and point-to-plane metrics are generally used for quality evaluation\cite{ref17}. Building on this foundation, \ADD{researchers} have proposed various quality assessment metrics by incorporating different features. Such metrics compare the different distance between points in three-dimensional space\cite{ref18}\cite{ref19}\cite{new1}, surface similarity\cite{ref20}, angle\cite{ref21}, curvature\cite{ref22}, color information statistics\cite{ref17} and other features to evaluate the point cloud quality. However, a common problem underlying is how to find the correspondence for a point in the distorted point cloud. \par
Unlike 2D images, 3D point clouds are unstructured data. Point clouds are not arranged in a regular grid. As a result, the location of a point is generally altered after compression, and in some cases, the point may vanish. This issue presents a challenge for full-reference quality assessment, which is essential for tasks such as compression and similar endeavors. When evaluating quality in full-reference, it is necessary to establish correspondences between points in the distorted point cloud and those in the original point cloud to assess the difference in distances or features at corresponding points. At present, the classical method to establish the correspondence between points relies on finding the spatial nearest neighbors. For example, when a point $A$ is missing at the same position in the distorted point cloud, its nearest neighbor is found as point $A'$, and the feature at $A'$ is used to approximate the feature at $A$. However, point clouds are nonuniform, with varying densities between points. The number and density of points can change dramatically after compression, leading to significant errors when approximating features at a point using those of its nearest neighbor.\par 
To address this issue, we propose a novel PCQA method that utilizes radial basis function (RBF) interpolation for accurate quality assessment by precisely correlating point cloud features.\par
The main contributions of this paper are as follows:
\begin{itemize}
\item{We propose to evaluate the feature difference for PCQA by building the bijective sets of point features for the original and distorted point clouds. The bijective feature sets ensure the one-to-one correspondence between the features at the points in the original and distorted point clouds. Such a design avoids the problem of many-to-one correspondence\ADD{,} which frequently happens when the point cloud density varies after compression. Moreover, this approach successfully bypasses the challenge where nearest neighbor searches fail to accurately capture neighborhood features when confronted with a substantial disparity in the number of points.}
\item{In order to build the bijective point feature sets, we employ the RBF for interpolation to convert discrete point features into a continuous feature function for the distorted point cloud. By substituting the geometry coordinates of the original point cloud into the feature function, we obtain the bijective sets of point features. PCQA is then performed by evaluating the feature difference between the corresponding points in the distorted and original point clouds. The proposed PCQA metric is named RBF-interpolation metric (RBFIM). The unique aspect of RBFIM is that the feature functions use point coordinates as the input and the corresponding point features as the output. This allows for the extraction of features at flexible coordinate positions. By using continuous feature functions, the features of the original and distorted point clouds are aligned.}
\item{As a result, RBFIM provides outstanding performance in PCQA. Compared with classic and state-of-the-art PCQA metrics, the proposed metric shows reliable performance on compression distortion in publicly accessible datasets.}
\end{itemize}
The paper is structured as follows: Section \ref{sec2} reviews previous work in the field. Section \ref{sec3} outlines the formulation of the problem and the fundamental principles behind RBFIM. The details of RBFIM are described in Section \ref{sec4}. Section \ref{sec6} introduces experimental results. Finally, conclusions are drawn in Section \ref{sec7}.\par
\section{Related Work}\label{sec2}
Existing full-reference PCQA algorithms can be broadly classified into two types based on the different information dimensions utilized in objective metrics: PCQA based on two-dimensional projection and PCQA using three-dimensional spatial features.\par
\subsection{PCQA based on Two-dimensional Projection }
The assessment of point cloud quality \ADD{projects the} reference and distorted point clouds onto a two-dimensional plane, \ADD{respectively,} followed by the evaluation of the resulting projection images using image quality assessment methods\ADD{. Such an approach} indirectly achieves the quality of \ADD{the} point cloud in 3D  assessment through the quality of images in 2D. Alexiou et al\cite{ref23} utilized various 3D visualization devices to validate the consistency between \ADD{the result of} 3D subjective quality assessment and \ADD{that of 2D projection,} proving that quality assessment consistent with subjective perception can be attained through appropriate plane projection. Torlig et al.\cite{ref24} proposed a PCQA algorithm utilizing bounding box plane projection. \ADD{Through} orthogonally projecting the point cloud onto the bounding box, six projection images were generated and evaluated for quality using image quality assessment algorithms such as PSNR, SSIM, and VIFP. The average quality of these projection images was then utilized to determine the overall quality of the point cloud. \ADD{Then} Alexiou et al.\cite{ref25} further investigated the impact of varying numbers of projection planes on the quality assessment results. By projecting the point cloud onto \ADD{different} planes \ADD{with} different solid shapes, they analyzed the correlation between objective evaluation and subjective \ADD{quality}. Their findings indicated that a 6-plane projection \ADD{can} effectively reflect \ADD{the} subjective quality of three-dimensional point clouds. Yang et al.\cite{ref26} extracted features from the planar projection images of the point cloud and developed a quality evaluation algorithm based on feature weighting, considering the impact of distortion on features. Similarly, Javaheri et al.\cite{ref27} adjusted the quality scores of projection images using geometric information from the point cloud, proposing a joint evaluation method based on geometry and attributes. Liu et al.\cite{ref28} proposed a structural similarity objective indicator based on attention mechanism and information content weighting to assess the quality of distorted point clouds. Freitas et al.\cite{ref29} projected point clouds as two-dimensional manifolds in three-dimensional space and mapped the attributes onto a folded two-dimensional grid to generate texture-containing images. \ADD{Then point} cloud quality assessment was conducted using image quality assessment. \par
Although resorting to available image quality assessment models is convenient and effective, PCQA based on two-dimensional projection often encounters issues of artefacts such as occlusions and voids, which affects the accuracy of quality evaluation. The computational complexity in projection is also prohibiting the application of related methods in practice.\par
\subsection{PCQA Using Three-dimensional Spatial Features}
Algorithms leveraging three-dimensional spatial features for PCQA are extensively utilized in practical applications. A typical example is the point-to-point distance-based metric used in the MPEG point cloud compression standard\cite{ref17}, which establishes correspondences between points in the distorted and reference clouds to measure the geometry and attribute errors. To enhance the measurement of point cloud geometry quality, Javaheri et al.\cite{ref18} replaced the Euclidean distance in the point-to-point distance-based metric with the Mahalanobis Distance, better reflecting spatial distribution. Tian et al.\cite{ref19} introduced a method for measuring geometry distortion based on point-to-plane distance, which quantifies geometry distortion by evaluating the projection distance along the normal direction between corresponding points in the distorted and original point clouds. Paolo et al.\cite{ref20} proposed meshing the original point cloud and computing the distance from each point in the distorted point cloud to the corresponding mesh to measure \ADD{geometry distortion}. Alexiou et al.\cite{ref21} suggested using the angle between the normals of corresponding points in the distorted and original point clouds to measure geometry distortion. Similarly, Meynet et al.\cite{ref22} utilized the disparity in curvature among neighboring surfaces of corresponding point pairs as a metric for distortion. They introduced a method for assessing geometry distortion in point clouds based on curvature. These algorithms offer the advantage of constructing distortion assessment models with low complexity\cite{ref30}. However, due to the limited consideration of features and the correspondence of points within the point clouds, their evaluation results often deviate significantly from the actual quality perceived by human observers.\par
To improve the performance of PCQA, researchers have endeavored to incorporate multiscale features to better capture the features of human perception. Building upon existing evaluation algorithms, Meynet et al.\cite{ref31} introduced the point cloud quality metric (PCQM), which integrates geometry and color information to approximate human-perceived quality. Specifically, PCQM extracts features such as curvature and luminance from the point cloud and combines them linearly to derive feature parameters closely resembling human perception. Similarly, Viola et al.\cite{ref32} extracted attribute features from statistical histograms and correlation graphs of point cloud attributes, combined with geometry local planar features, to obtain features that characterize the overall visual quality of the point cloud. Yang et al.\cite{ref33} extracted key points from three-dimensional point clouds and established a multiscale \ADD{method for joint quality evaluation of geometry and attribute} using gradient features of local point cloud graphs constructed from crucial points\cite{ref34}. Alexiou et al.\cite{ref35} proposed an objective quality evaluation method by studying the geometry topological relationships and local distribution of color attributes in point clouds. Wang et al.\cite{ref36} proposed a method for joint quality evaluation of geometry and attributes by utilizing multiscale texture features from 2D images and geometry features from 3D points. Lu et al.\cite{ref37} utilized a dual-scale 3D Gaussian difference filter to extract 3D edge features and evaluated point cloud quality using 3D edge similarity. Yang et al.\cite{ref38} measured distortions by comparing the differences using multiscale potential energy discrepancy between distorted and original point clouds. Zhang et al.\cite{ref39} measured distortions using transformational complexity. Although these methods achieve improved \ADD{performance in evaluation} through fitting or learning model parameters, they face a common problem of relating the correspondence for a point in the distorted point cloud.\par
\section{Problem Formulation}\label{sec3}
\begin{figure*}[!t]
\centering
\subfloat[\ADD{searching nearest neighbor}]{\includegraphics[width=6.7in]{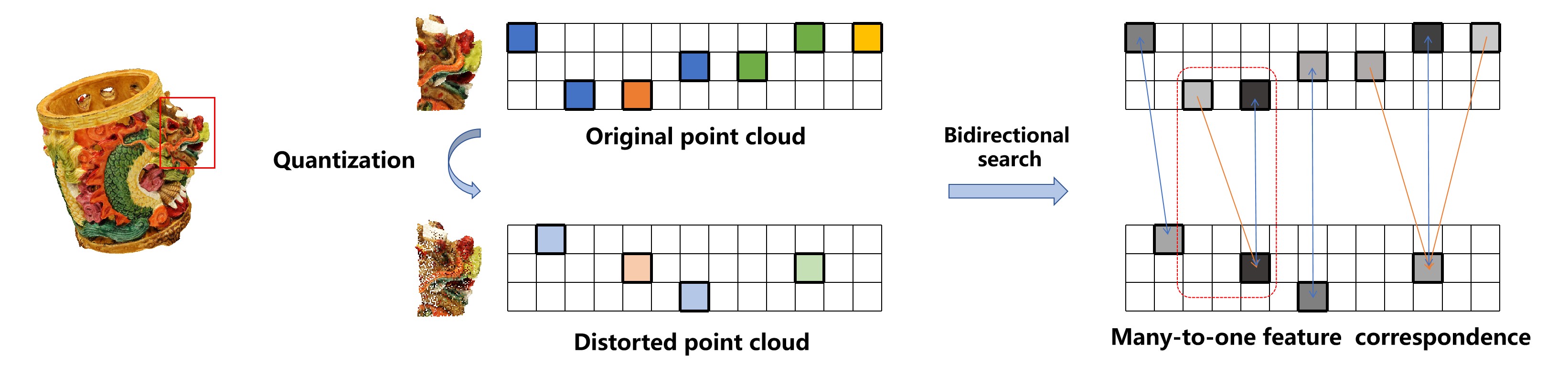}}%
\label{fig1_a}
\hfil
\subfloat[\ADD{RBFIM}]{\includegraphics[width=6.5in]{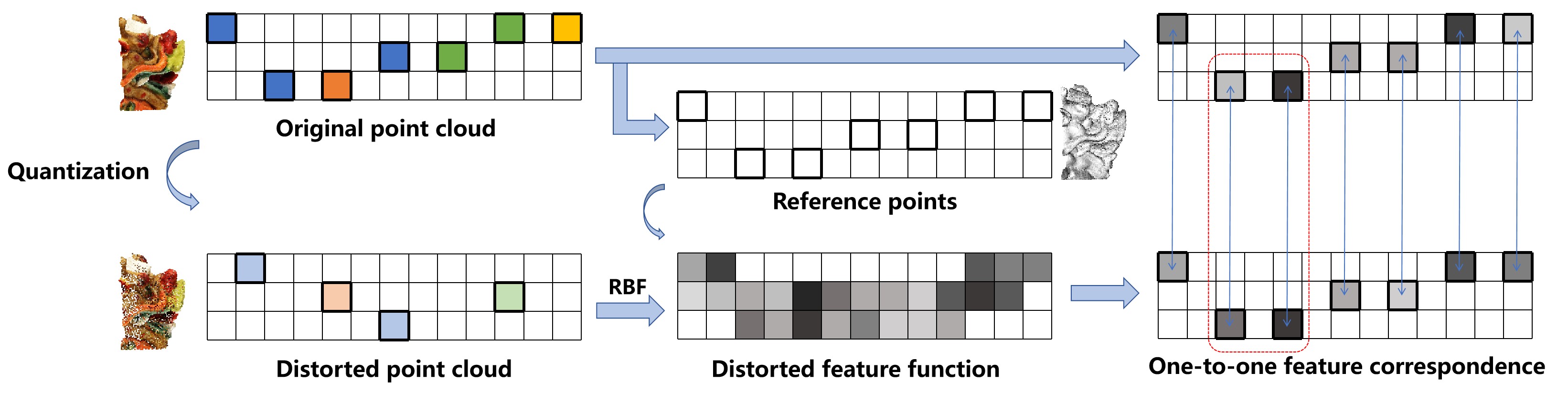}}%
\label{fig1_b}
\caption{\ADD{Comparison of distortion calculations between different methods.  $\to$ represents unidirectional mapping, $\leftrightarrow $ represents bidirectional mapping.}}
\label{fig_1}
\end{figure*}
A point cloud is essentially a collection of data points sampled from the surface of a 3D model. Let ${{\bf{P}}_O} = \{ {{\bf{p}}_i} \in {{\Bbb{R}}^3}\} _{i = 1}^{{N_O}}$ be an original point cloud with ${N_O}$ points and ${{\bf{P}}_D} = \{ {{\bf{p}}_j} \in {\Bbb{R}}{^3}\} _{j = 1}^{{N_D}}$ be a distorted point cloud of ${{\bf{P}}_O}$ with ${N_D}$ points. The perceptual distortion between these point clouds, ${{\bf{P}}_O}$ and ${{\bf{P}}_D}$, is measured by $D({{\bf{P}}_O},{{\bf{P}}_D})$, as defined by\par
\begin{equation}
\left\{ \begin{array}{l}
D = {\raise0.7ex\hbox{$1$} \!\mathord{\left/
 {\vphantom {1 {{N_O}}}}\right.\kern-\nulldelimiterspace}
\!\lower0.7ex\hbox{${{N_O}}$}} \cdot \sum\limits_{i = 1}^{{N_O}} {{d_i}} \\
{d_i} = \left|{{{\text{M}}_O}({{\bf{p}}_i}) - {{\text{M}}_D}({{{\bf{p'}}}_i})}\right|
\end{array} \right.\text{.}
\label{eq1}
\end{equation}
Here, $d_i$ represents the difference in perceptual features for a point ${\bf{p}}_i$.  ${\text{M}}_O({\bf{p}}_i)$ denotes the perceptual feature of ${\bf{p}}_i$ in ${\bf{P}}_O$, while ${\text{M}}_D({\bf{p'}}_i)$ denotes the perceptual feature of a point ${\bf{p'}}_i$ in ${\bf{P}}_D$. ${\bf{p}}_i$ is a point in ${\bf{P}}_O$, and ${\bf{p'}}_i$ is its corresponding point in ${\bf{P}}_D$. The absolute value is represented by $\left|. \right|$. Therefore, accurate correspondence of features between the original and distorted point clouds is a critical factor affecting the performance of PCQA.\par
The correspondence of features depends on the position of points, and a common challenge underlying is establishing a proper correspondence between points in ${\bf{P}}_O$ and ${\bf{P}}_D$. This difficulty arises due to variations in the position and number of points between the two point clouds, making it \ADD{generally not possible} to establish a one-to-one correspondence similar to \ADD{pixel-level alignment} in image quality assessments.\par
A classic approach to \ADD{find the correspondence for a point cloud is to use} the searching nearest neighbor method. The nearest point searched in ${\bf{P}}_D$ to a point ${\bf{p}}_i$ in ${\bf{P}}_O$ is chosen as the corresponding point, and the nearest point's feature is used to calculate the perceptual feature difference. As shown in 
\begin{equation}
{{d'}_i} =  \left|{{{\text{M}}_O}({{\bf{p}}_i}) - {{\text{M}}_D}({{\bf{p''}}_i})}\right|\text{,}
\label{eq2}
\end{equation}
${{\bf{p''}}_i}$ is the nearest neighbor of ${\bf{p}}_i$ in ${{\bf{P}}_D}$. \par
\ADD{However,} many-to-one correspondence often arises when the number of points changes due to compression \ADD{or processing, where} the classic method \ADD{generally} requires bidirectional searching \ADD{as shown in Fig.\ref{fig_1} (a)}. And the final distortion ${D_{classic}}$ calculated by the searching nearest neighbor method is described by
\begin{equation}
{D_{classic}} = \max \{ {D_1},{D_2}\} 
\label{eq3} \ADD{\text{,}}
\end{equation}
\ADD{where} ${D_1}$ is the distortion from ${{\bf{P}}_O}$ to ${{\bf{P}}_D}$, and ${D_2}$ is the distortion from ${{\bf{P}}_D}$ to ${{\bf{P}}_O}$.\par 
The classic method uses the features of the nearest neighbor points to calculate ${{d'}_i}$, as spatially closer points are likely to exhibit similar perceptual features. The similarity is highly dependent on the spatial distance between points and their nearest neighbors - the shorter the distance, the higher the similarity, and vice versa. Consequently, larger distances can result in significant inaccuracies in the calculated distortions ${d'}_i$, which impact the accuracy of ${D_{classic}}$. However, it is not uncommon for PCC to involve larger \ADD{average Euclidean} distance in corresponding points, particularly in G-PCC, as shown in Table \ref{qualtized}. Multiple points are often merged using octree pruning, which leads to substantial changes in both the number and position of points within the quantized point clouds. As illustrated in Fig.\ref{fig_1} (a), in the red dashed box, due to changes in the position and number of points caused by PCC, the blue point in ${{\bf{P}}_O}$ is paired with the pink point in ${{\bf{P}}_D}$. In this case, the calculated ${d'}_i$ is different from the actual perceived feature difference. On the other hand, the bidirectional search has high computational complexity.\par 
\begin{table}[]
\centering
\setlength{\tabcolsep}{4.0pt}
 \caption{Comparison of the number and position of points between the quantized and the original point cloud in G-PCC}
 \begin{adjustbox}{max width=\textwidth}
 \renewcommand{\arraystretch}{1.2}
\begin{tabular}{ccccc}
\hline
\multirow{2}{*}{Sequences}                                                   & \multicolumn{2}{c}{The number of points} & \multirow{2}{*}{\begin{tabular}[c]{@{}c@{}}Geometry\\ quantization\end{tabular}} & \multirow{2}{*}{\ADD{\begin{tabular}[c]{@{}c@{}}Average Euclidean\\ distance\end{tabular}}} \\ \cline{2-3}
                                                                             & original           & quantized           &                                                                               &                           \\ \hline
\multirow{4}{*}{longdress}                                                   & 857966             & 4092                & 1/32                                                                          & 14.89                     \\
                                                                             & 857966             & 15688               & 1/16                                                                          & 7.27                      \\
                                                                             & 857966             & 62130               & 1/8                                                                           & 3.75                      \\
                                                                             & 857966             & 238492              & 1/4                                                                           & 1.85                      \\ \hline
\multirow{4}{*}{\begin{tabular}[c]{@{}c@{}}basketball\\ player\end{tabular}} & 2925514            & 5099                & 1/32                                                                          & 14.91                     \\
                                                                             & 2925514            & 13209               & 1/16                                                                          & 7.26                      \\
                                                                             & 2925514            & 52468               & 1/8                                                                           & 3.79                      \\
                                                                             & 2925514            & 206712              & 1/4                                                                           & 1.93                      \\ \hline
\end{tabular}
\end{adjustbox}
\label{qualtized}
\end{table}
To address the problem described \ADD{above}, we propose a method called RBFIM. We adopt RBF interpolation to transform the discrete features of points in ${{\bf{P}}_D}$ into a continuous feature function, and using points in ${{\bf{P}}_O}$ as reference points. By substituting the coordinates of reference points into this continuous feature function, we are able to generate bijective sets of point features, effectively mapping each feature in ${{\bf{P}}_O}$ to a corresponding feature in ${{\bf{P}}_D}$. Then, the distortion between ${{\bf{P}}_D}$ and ${{\bf{P}}_O}$ is calculated by bijective feature sets, as shown in
\begin{equation}
\left\{ \begin{array}{l}
{D_{proposed}} = {\raise0.7ex\hbox{$1$} \!\mathord{\left/
 {\vphantom {1 {{N_R}}}}\right.\kern-\nulldelimiterspace}
\!\lower0.7ex\hbox{${{N_R}}$}} \cdot \sum\limits_{k = 1}^{{N_R}} {{d''_{k}}} \\
{d''_{k}} = \left|{{\text{M}_O}({{\bf{p}}_k}) - {f_D}({{\bf{p}}_k})}\right|
\end{array} \right.\text{,}
\label{eq4}
\end{equation}
where ${{\bf{P}}_R} = \left\{ {{{\bf{p}}_k} \in {{\Bbb{R}}^3}} \right\}_{k = 1}^{{N_R}}$ represents a set of reference points and ${{\bf{P}}_R} \subseteq {{\bf{P}}_O}$. ${f_D}({\bf{p}})$ is the continuous feature function calculated by ${{\text{M}}_D}({{\bf{p}}_j})$ using RBF interpolation. \par
As illustrated in Fig. \ref{fig_1} (b), features correspond one to one \ADD{using RBFIM}. In the red dashed box, the feature on the blue point in ${{\bf{P}}_O}$ corresponds to the feature calculated by the feature function at the same coordinates in ${{\bf{P}}_D}$. Therefore, RBFIM allows for a more \ADD{reasonable and flexible way of finding the} correspondence of features \ADD{when} the number and position of points \ADD{differ in two point clouds}. And the feature function for ${{\bf{P}}_D}$ is constructed instead of ${{\bf{P}}_O}$ \ADD{considering} the complexity of interpolation \ADD{since the} number of points of the original point cloud is generally larger than that of the compressed point cloud. \par
\section{Proposed Method}\label{sec4}
\begin{figure*}[h]
\centering  
\includegraphics[width=6.0in]{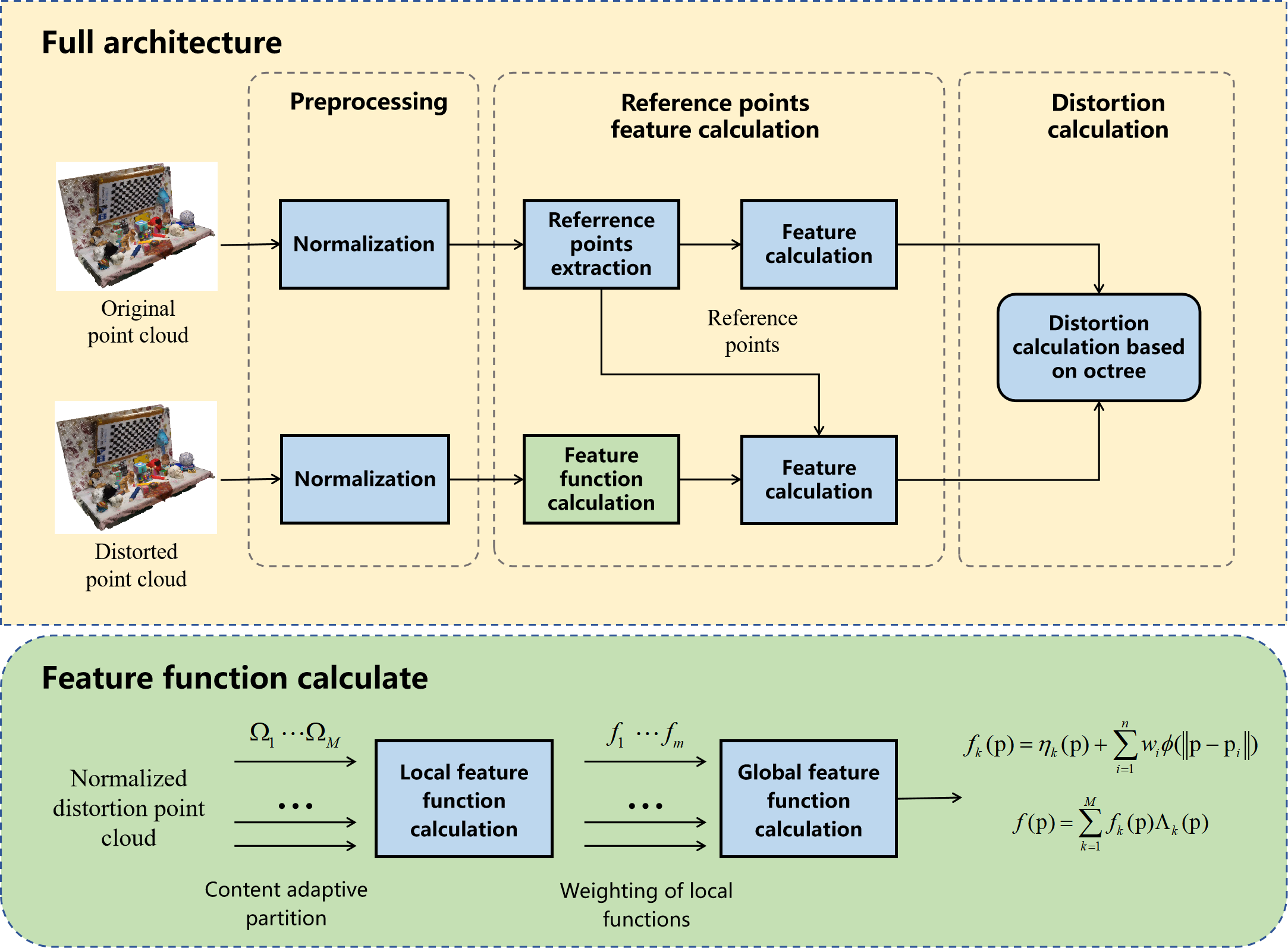}
\caption{The schematic diagram depicts the process of implementing the RBFIM solution. The feature function calculation involves both local and global feature functions. $\Omega _1 \cdots \Omega _M$ are the point sets divided into the normalized distorted point cloud, $f_1 \cdots f_M$ are the local feature functions, and $f({\bf{p}})$ is the global feature function.}
\label{fig2}
\end{figure*}
In this section, we first present the general framework of the proposed RBFIM in subsection A. Following that, the implementation details of each module are described in subsections B-D, respectively. 
\subsection{Overview}
Fig. \ref{fig2} shows the general framework of the RBFIM method. The first step is preprocessing, which takes the original point cloud and distorted point cloud as input, and then normalizes them. The second step is to calculate the feature of the distorted point cloud at the position of reference points which are extracted from the normalized original point cloud. In this work, we use RBF to interpolate the feature function of the normalized distorted point cloud. The features at corresponding points in the distorted point cloud are obtained by substituting the coordinates of reference points into the feature function of the distorted point cloud. The calculation of the feature function involves both local and global feature functions, as described in Section IV.C. The last step is to calculate the distortion through the feature differences at the reference points to get the final quality for the distorted point cloud.\par

\subsection{Preprocessing}
To eliminate the impact of different geometry scales on objective PCQA and to generalize the evaluation method for different geometry bit-widths, we first normalize both the distorted and the original point cloud before subsequent processing.\par 
${{\bf{P}}_O} = \{ {{\bf{p}}_i} \in {{\Bbb{R}}^3}\} _{i = 1}^{{N_O}}$ is an original point cloud and ${{\bf{P}}_D} = \{ {{\bf{p}}_j} \in {\Bbb{R}}{^3}\} _{j = 1}^{{N_D}}$ is a distorted point cloud of ${{\bf{P}}_O}$, where each ${{\bf{p}}_i}, {{\bf{p}}_j} \in {{\Bbb{R}}^3}$ is a vector with a 3D coordinate (${\bf{p}} = (x,y,z)$). ${{\bf{\hat P}}_O}$ and ${{{\bf{\hat P}}}_D}$ denote the normalized point clouds corresponding to $\mathbf{P}_O$ and $\mathbf{P}_D$, respectively. ${{{\bf{\hat p}}}_i}$ and ${{{\bf{\hat p}}}_j}$ are the points of ${{{\bf{p}}}_i}$ and ${{{\bf{p}}}_j}$ after normalization, where ${{{\bf{\hat p}}}_i} \in {{\bf{\hat P}}_O}$ and ${{{\bf{\hat p}}}_j} \in {{\bf{\hat P}}_D}$. The normalization is described in
\begin{equation}
\ADD{
\left\{ \begin{array}{l}
{{{\bf{\hat p}}}_i} = {{1024 \cdot \left( {{{\bf{p}}_i} - {{\bf{p}}_{\min }}} \right)} \mathord{\left/
 {\vphantom {{1024 \cdot \left( {{{\bf{p}}_i} - {{\bf{p}}_{\min }}} \right)} {{L_{max}}}}} \right.
 \kern-\nulldelimiterspace} {{L_{max}}}}\\
{{{\bf{\hat p}}}_j} = {{1024 \cdot \left( {{{\bf{p}}_j} - {{\bf{p}}_{\min }}} \right)} \mathord{\left/
 {\vphantom {{1024 \cdot \left( {{{\bf{p}}_j} - {{\bf{p}}_{\min }}} \right)} {{L_{max}}}}} \right.
 \kern-\nulldelimiterspace} {{L_{max}}}}
\end{array} \right.
}
\text{,}
\label{normalized}
\end{equation}
where $L_{max}$ is the length of the longest edge of the bounding box of ${{\bf{P}}_O}$ while the point ${{\bf{p}}_{\min }} = ({x_{\min}},{y_{\min}},{z_{\min}})$ is composed of the minimum value of the coordinate at $x, y, z$ directions, respectively.\par
\subsection{Feature Calculation for Distorted Point Cloud at Reference Points}
Considering the complexity and the relevance of \ADD{the} selected feature to human visual perception, we use luminance as \ADD{the feature for a point. Indeed}, other features can also be selected, and we verify the effects of different features in the experimental section. \CZADD{The proposed method can still be used for PCQA even when the point cloud lacks color information. In such a case, geometry features only, such as curvature, can be used. The corresponding performance of the proposed method using curvature only will be evaluated in Section V. E.}\par
Let ${I_O}({{\bf{\hat p}}_i}) = \{ {v_i}\} _{i = 1}^{{N_O}}$ represent the luminance of ${\bf{\hat{P}}}_O$, and ${I_D}({{\bf{\hat p}}_j}) = \{ {v_j}\} _{j = 1}^{{N_D}}$ denote the luminance of ${\bf{\hat{P}}}_D$. To obtain the luminance at the reference points, we first calculate the feature function of ${\bf{\hat{P}}}_D$. According to the RBF interpolation principle\cite{ref40}, since the feature function passes through the given set of points and has high smoothness, the problem of solving the feature function can generally be transformed into finding a function $f({\bf{p}})$ that minimizes the variational function $V$, as shown in 
\begin{equation}
\arg \min {V}[f({{{\bf{\hat p}}}_j})] = \arg \min \left\{ {\sum\limits_{j = 1}^{{N_D}} {{{\left[ {f({{{\bf{\hat p}}}_j}) - {v_j}} \right]}^2}}  + \lambda E[f]} \right\}
\label{RBF-principle} \text{.}
\end{equation}
The first term on the right side of Eq. (\ref{RBF-principle}) ensures the accuracy of the feature function, while the second term $E[f]$ ensures its smoothness, and $\lambda  > 0$ is the regularization parameter used to find the optimal balance between accuracy and smoothness. \ADD{$f({\bf{p}})$ can usually be simplified to \cite{ref40}}
\begin{equation}
f({\bf{p}}) = \eta ({\bf{p}}) + \sum\limits_{j = 1}^{{N_D}} {{\omega _j}\phi \left( {{{\left\| {{\bf{p}} - {{{\bf{\hat p}}}_j}} \right\|}_2}} \right)}
\label{RBF} \text{.}
\end{equation}
Eq. (\ref{RBF}) represents the general formula of RBF, where $\eta ({\bf{p}})$ is a set of bases in three dimensions. $\eta ({\bf{p}})$ is a three-variable polynomial with a maximum degree of 3. It is commonly expressed as $\eta ({\bf{p}}) = {\text{a}}x +{\text{b}}y+{\text{c}}z+\text{d}$, where ${\text{a,b,c,d}}$ are constant coefficients. ${\omega _i}$ denote the weight coefficients, and ${\left\|  \cdot  \right\|_2}$ denotes the Euclidean norm. To ensure orthogonality, the weight coefficients must satisfy the following constraint conditions:
\begin{equation}
\sum\limits_{j = 1}^{N_D} {{\omega _j}}  = \sum\limits_{j = 1}^{N_D} {{\omega _j}} {\hat{x}}_j = \sum\limits_{j = 1}^{N_D} {{\omega _j}} {\hat{y}}_j = \sum\limits_{j = 1}^{N_D} {{\omega _j}} {\hat{z}}_j = 0
\label{constraint}\text{.}
\end{equation}
Here, ${\hat{x}}_j$, ${\hat{y}}_j$, and ${\hat{z}}_j$ represent coordinates of point ${{{\bf{\hat p}}}_j}$ at $x, y, z$ directions, respectively. And by inputting all coordinates of ${{{\bf{\hat p}}}_j}$ in ${{\bf{\hat P}}_D}$ into Eq. (\ref{RBF}), we determine the coefficients of $\eta ({\bf{p}})$ and ${\omega _i}$ through the following equation:  
\begin{equation}
{\mathbf{X}} \cdot {\mathbf{W}} = {\mathbf{Y}}
\label{matrix} \text{.}
\end{equation} 
In Eq. (\ref{matrix}), ${\mathbf{Y}}$ is the feature matrix, as shown in 
\begin{equation}
{\mathbf{Y}} = {\left[\!\!\!\!{\begin{array}{*{20}{c}}
{\begin{array}{*{20}{c}}\! 
{{I_D}({{{\bf{\hat p}}}_1})}&{{I_D}({{{\bf{\hat p}}}_2})}& \cdots &{{I_D}({{{\bf{\hat p}}}_{{N_D}}})}
\end{array}}&{\begin{array}{*{20}{c}}\!\!\!\!\!\!\!\!
{\text{0}}&{\text{0}}&{\text{0}}&{\text{0}}
\end{array}}
\end{array}} \!\!\!\!\right]^ \top}
\label{matrixY}\text{.}
\end{equation} 
${\mathbf{X}}$ is the coordinate matrix, as represented in 
 \begin{equation}
{\mathbf{X}} = \left[\!\!\!\!{\begin{array}{*{20}{c}}
{{\phi _{11}}}&{{\phi _{12}}}& \cdots &{{\phi _{1{N_D}}}}&1&{{{\hat x}_1}}&{{{\hat y}_1}}&{{{\hat z}_1}}\\
{{\phi _{21}}}&{{\phi _{22}}}& \cdots &{{\phi _{2{N_D}}}}&1&{{{\hat x}_2}}&{{{\hat y}_2}}&{{{\hat z}_2}}\\
 \vdots & \vdots & \ddots & \vdots & \vdots & \vdots & \vdots & \vdots \\
{{\phi _{{N_D}1}}}&{{\phi _{{N_D}2}}}& \cdots &{{\phi _{{N_D}{N_D}}}}&1&{{{\hat x}_{{N_D}}}}&{{{\hat y}_{{N_D}}}}&{{{\hat z}_{{N_D}}}}\\
1&1& \cdots &1&0&0&0&0\\
{{{\hat x}_1}}&{{{\hat x}_2}}& \cdots &{{{\hat x}_{{N_D}}}}&0&0&0&0\\
{{{\hat y}_1}}&{{{\hat y}_2}}& \cdots &{{{\hat y}_{{N_D}}}}&0&0&0&0\\
{{{\hat z}_1}}&{{{\hat z}_2}}& \cdots &{{{\hat z}_{{N_D}}}}&0&0&0&0
\end{array}}\!\!\!\!\!\right]\text{,}
\label{matrixX}
\end{equation}
where ${{\phi _{21}}}$ is equal to ${\phi \left( {{{\left\| {{\bf{\hat p}}_2 - {{{\bf{\hat p}}}_1}} \right\|}_2}} \right)}$. And ${\mathbf{W}}$ is the weight matrix, as represented in 
\begin{equation}
{\mathbf{W}} = {\left[\!\!\!\!{\begin{array}{*{20}{c}}
{\begin{array}{*{20}{c}}
{{\omega _1}}&{{\omega _2}}& \cdots &{{\omega _{{N_D}}}}
\end{array}}&{\begin{array}{*{20}{c}}\!\!\!\!\!\!\!\!
{\text{a}}&{\text{b}}&{\text{c}}&{\text{d}}
\end{array}}
\end{array}} \!\!\!\!\right]^ \top }
\label{matrixW}\text{.}
\end{equation}  
Based on the solution of Eq. (\ref{matrix}), the feature function $f({\bf{p}})$ is obtained. By substituting the reference points set ${{\bf{P}}_R} \subseteq {{\bf{\hat P}}_O}$ into $f({\bf{p}})$, luminance of the normalized distorted point cloud corresponding to ${{\bf{P}}_R}$ are obtained. And luminance of the normalized original point cloud in ${{\bf{P}}_R}$ can be represented by ${I_O}({{\bf{\hat p}}_i}) = \{{v_i}\} _{i = 1}^{{N_O}}$.\par
However, in practical calculations, the cost of solving RBF parameters for the entire point cloud is enormous, \ADD{because the number of linear equations is determined by the number of points} \cite{ref41}. To improve algorithm efficiency while ensuring the accuracy of the feature function, we \ADD{use} a piece-wise approach to compute the feature function. First, using the adaptive spatial partitioning method in Multi-level Partition of Unity implicits\cite{ref42}, ${{\bf{\hat P}}_D}$ is divided into multiple overlapping node subdomains. Then, the local feature function for each node subdomain is computed separately, and the global feature function is obtained by weighting and summing the local feature functions. This approach reduces computational complexity while ensuring the accuracy of the feature function description. The specific technical details are as follows.\par
\begin{itemize}
\item{Local feature functions calculation}
\end{itemize}
An adaptive partition method is used to divide ${{\bf{\hat P}}_D}$ into a number of smaller octree nodes covering each other sparsely.\par
The algorithm starts by setting a local support radius $R$ for each octree node, which is the length of \ADD{the central diagonal of the bounding box}. Each node is assigned to contain a specific number of sample points within a range defined by a minimum value, ${T_{\min}}$, and a maximum value ${T_{\max}}$. To control the depth of subdivision, we introduce the maximum norm of the local approximation error, denoted as $\varepsilon $, based on the Taubin distance\cite{ref42}. As shown in 
\begin{equation}
\varepsilon  = \mathop {\max }\limits_{{{\left\| {{{{\bf{\hat p}}}_j} - {{\bf{c}}_j}} \right\|}_2} < R} \frac{{\left| {G({{{\bf{\hat p}}}_j})} \right|}}{{\left| {\nabla G({{{\bf{\hat p}}}_j})} \right|}}
\label{Taubin}\text{,}
\end{equation}
the gradient of $G({{\bf{\hat{p}}}_j})$, denoted as $\nabla G({{\bf{\hat{p}}}_j})$, is calculated, and the center of the node is represented by ${{\bf{c}}_j}$. When each level of node ${C^{[l]}}$ is generated, we approximate the local shape function $G({{\bf{\hat{p}}}_j})$ using the fast least squares method, followed by calculating $\varepsilon$. If $\varepsilon$ exceeds the specified error threshold ${\varepsilon _0}$, subdivision continues. To improve computational efficiency, computation begins from $l \ge 4$ ($l$ represents the level of octree partition).

\begin{algorithm}
\caption{Octree-decompose(${\bf{\hat{P}}}_D$, ${C^{[l]}}$, $l$)} 
\label{alg1}
\begin{algorithmic}
\REQUIRE Point cloud ${\bf{\hat{P}}}_D$, cubic cell ${C^{[l]}}$, level $l$ 
\ENSURE The set of subdomains ${\bf{D}} = \{\Omega _{k}\} _{k = 1}^M$
\STATE Compute $\Omega _{k}^{[l]}$ with $R$ 
\STATE Set $n$ the number of points in $\Omega _{k}^{[l]}$
\IF{$n > {T_{\max }}$} 
\IF{$l < 4$}
\STATE Subdivide $C_{}^{^{[l]}}$ into cubic octant cells $C_1^{^{[l + 1]}},...,C_8^{^{[l + 1]}}$
\STATE Octree-decompose($\Omega $, $C_1^{^{[l + 1]}}$, $l + 1$)
\STATE ...
\STATE Octree-decompose($\Omega $, $C_8^{^{[l + 1]}}$, $l + 1$)
\ELSIF{$l >  = 4$}
\STATE Compute ${\varepsilon _{}}$ with $\max (\left| {G({{\bf{\hat{p}}}_j})} \right|/\left| {\nabla G({{\bf{\hat{p}}}_j})} \right|)$
\IF{$\varepsilon  > {\varepsilon _0}$}
\STATE Subdivide $C_{}^{^{[l]}}$ into cubic octant cells $C_1^{^{[l + 1]}},...,C_8^{^{[l + 1]}}$
\STATE Octree-decompose($\Omega $, $C_1^{^{[l + 1]}}$, $l + 1$)
\STATE ...
\STATE Octree-decompose($\Omega $, $C_8^{^{[l + 1]}}$, $l + 1$)
\ELSE 
\STATE Stop subdivision
\ENDIF
\STATE Domain is OK, add $\Omega _k^{[l]}$ to $\bf{D}$
\ENDIF
\STATE Domain is OK, add $\Omega _k^{[l]}$ to $\bf{D}$
\ELSIF{$n < {T_{\min}}$}
\WHILE{$n \notin [{T_{\min }},{T_{\max }}]$} 
\IF{$n < {T_{\min }}$} 
\STATE Enlarge $\Omega _k^{[l]}$ 
\ELSIF{$n > {T_{\max }}$}
\STATE Decrease $\Omega _k^{[l]}$
\ENDIF
\STATE Set $n$ the number of points in $\Omega _k^{[l]}$ 
\ENDWHILE
\STATE Domain is OK, add $\Omega _k^{[l]}$ to $\bf{D}$
\ELSE
\STATE Domain is OK, add $\Omega _k^{[l]}$ to $\bf{D}$
\ENDIF
\end{algorithmic}
\end{algorithm}

To compute the local RBF function, each node needs to have a sufficient number of sample points. If the number of sample points contained in node ${C^{[l]}}$ are less than ${T_{\min}}$, it is necessary to increase $R$. And if the number of samples exceeds ${T_{\max}}$, the node is further subdivided. Additionally, if the node unit is empty, it should be removed. Algorithm \ref{alg1} provides the pseudo-code for the adaptive octree partition algorithm. The normalized point cloud ${\bf{\hat{P}}}_D$ is the global domain. After adaptive partition, ${\bf{\hat{P}}}_D$ is divided into $M$ mutually overlapping spatial subdomains ${\{{\Omega _k}\} _{k \in [1,M]}}$, and ${{\bf{\hat{P}}}_D} \supseteq \bigcup\limits_k {{\Omega _k}} $. \par
For the subdomain ${\{ {\Omega _k}\} _{k \in [1,M]}}$, the corresponding local feature function ${\{ {{{f_k}({\bf{p}})}\} _{k \in [1,M]}}}$ can be obtained by Eq. (\ref{matrix}).\par
\begin{itemize}
\item{Global feature function calculation}
\end{itemize}
Subsequently, to obtain the global function $f({\bf{p}})$, we use non-negative blending functions ${\{ {\Lambda _k}\} _{k \in [1,M]}}$ to weight the functions $f_k({\bf{p}})$\cite{ref45}. The calculation formula is shown in
\begin{equation}
f({\bf{p}}) = \sum\limits_{k = 1}^M {{f_k}({\bf{p}}){\Lambda _k}({\bf{p}})} 
\label{weight}\text{.}
\end{equation}
The blending function ${\{ {\Lambda _k}\} _{k \in [1,~M]}}$ has finite support ${\text{sup}}\{ {\Lambda _k}\}  \subseteq {\Omega _k}$, denoted as $\forall {\bf{p}} \notin {\Omega _k} \Rightarrow {\Lambda _k} = 0$, and the sum over the entire domain equals 1, that is: $\sum\limits_{k = 1}^M {{\Lambda _k}({{\bf{\hat{p}}}_j})}  = 1,{{\bf{\hat{p}}}_j} \in {\bf{\hat{P}}}_D$. And ${\Lambda _k}$ can be obtained from 
\begin{equation}
{\Lambda _k}({\bf{p}}) = \frac{{{w_k}({\bf{p}})}}{{\sum\limits_{k = 1}^M {{w_k}({\bf{p}})} }}
\label{Lambda}\text{,}
\end{equation}
\ADD{where }${{w_k}({\bf{p}})}$ representing the weight function.\par
In order to improve the computational efficiency, we simplify the calculation of ${{w_k}({\bf{p}})}$ by using the inverse distance weighting function, which is expressed as 
\begin{equation}
{w_k}({\bf{p}}) = {\left[ {\frac{{\max ({R_k} - \left\| {{\bf{p}} - {{\bf{c}}_k}} \right\|_2,0)}}{{{R_k} \cdot \left\| {{\bf{p}} - {{\bf{c}}_k}} \right\|_2}}} \right]^2}
\label{weight2}\text{.}
\end{equation}
Here ${R_k}$ represents the maximum radius of the subdomain ${\Omega _k}$, and ${{\bf{c}}_k}$ stands for the center of ${\Omega _k}$. The function ${w_k}$ is not only structurally simple but also compactly supported, meaning that $\left\| {{\bf{p}} - {{\bf{c}}_k}} \right\|_2 \ge {R_k} \Rightarrow {w_k}({\bf{p}}) = 0$.\par

\subsection{Distortion Calculation}
Following the above steps, luminance of ${\bf{\hat{P}}}_D$ and ${\bf{\hat{P}}}_O$ at all ${\bf{p}}_k \in {\bf{\hat{P}}}_R$ are obtained. Because \ADD{the luminance value at each point has} one-to-one correspondence, various methods can be used to calculate distortion. We choose the simplest method of calculating mean squared error (MSE) as distortion. Considering the human visual attention mechanism \cite{ref46}, an octree partitioning is used to divide ${{\bf{P}}_R}$ into ${M_R}$ nodes of equal size, \ADD{then, distortion in Eq. (\ref{eq4}) is} computed as 
\begin{equation}
{D_{\text{RBFIM}}} = \frac{1}{{{M_R}}}\sum\limits_{i = 1}^{{M_R}} {\left| {{\rm{Mean}}_i^D - {\rm{Mean}}_i^O} \right|} 
\label{mse} \text{,}
\end{equation}
where ${{\text{Mean}}_i^O}$ and ${{\text{Mean}}_i^D}$ are the average luminance of \ADD{${\bf{\hat{P}}}_O$ and ${\bf{\hat{P}}}_D$} in each node, \ADD{respectively. Referring to the distortion and quality mapping method\cite{ref17}, the perceptual quality of RBFIM is calculated as 
\begin{equation}
{Q_{{\rm{RBFIM}}}} = 20 \cdot {\rm{lo}}{{\rm{g}}_{10}}\left( {{{255} \mathord{\left/
 {\vphantom {{255} {{D_{{\rm{RBFIM}}}}}}} \right.
 \kern-\nulldelimiterspace} {{D_{{\rm{RBFIM}}}}}}} \right)
\label{psnr} \text{.}
\end{equation}}
\section{Experimental Evaluations}\label{sec6}
In this section, we utilize multiple point cloud datasets to validate the effectiveness of the proposed RBFIM. Furthermore, we compare the quality assessment results of RBFIM with classic and state-of-the-art (SOTA) PCQA metrics on G-PCC and V-PCC distorted point clouds.
\begin{table*}[tp]
\centering
\setlength{\tabcolsep}{2.0pt}
 \caption{Performance comparison of different classic and SOTA metrics on 5 datasets. The best and second-best are highlighted in \textcolor{red}{red} and \textcolor{blue}{blue}, respectively.}
 \begin{adjustbox}{max width=\textwidth}
\begin{tabular}{ccccccccccccccccccc}
\hline
\multicolumn{1}{c|}{}                                                                            &                         &                            & \multicolumn{5}{c|}{Multi-feature metrics}                                                                                                                                                                                                                                                                                                     & \multicolumn{10}{c|}{Single-feature metrics}                                                                                                                                                                                                                                                                                                                                                                                                                                                                                                                                                                                                      &                              \\ \cline{4-18}
\multicolumn{1}{c|}{\multirow{-2}{*}{Datasets}}                                                  & \multirow{-2}{*}{PCC}   & \multirow{-2}{*}{Criteria} & \begin{tabular}[c]{@{}c@{}}GraphSIM\\ \cite{ref33}\end{tabular} & \begin{tabular}[c]{@{}c@{}}MS\_GraphSIM\\ \cite{ref34}\end{tabular} & \begin{tabular}[c]{@{}c@{}}PCQM\\ \cite{ref31}\end{tabular} & \begin{tabular}[c]{@{}c@{}}${\text{pSSIM}}_\text{g}$\\ \cite{ref35}\end{tabular} & \multicolumn{1}{c|}{\begin{tabular}[c]{@{}c@{}}${\text{pSSIM}}_\text{c}$\\ \cite{ref35}\end{tabular}} & \begin{tabular}[c]{@{}c@{}}${\text{MSE}}_{p2po}$\\ \cite{ref17}\end{tabular} & \begin{tabular}[c]{@{}c@{}}${\text{PSNR}}_{p2po}$\\ \cite{ref17}\end{tabular} & \begin{tabular}[c]{@{}c@{}}${\text{MSE}}_{p2pl}$\\ \cite{ref17}\end{tabular} & \begin{tabular}[c]{@{}c@{}}${\text{PSNR}}_{p2pl}$\\ \cite{ref17}\end{tabular} & \begin{tabular}[c]{@{}c@{}}${\text{MSE}}_\text{Y}$\\ \cite{ref17}\end{tabular} & \begin{tabular}[c]{@{}c@{}}${\text{MSE}}_\text{U}$\\ \cite{ref17}\end{tabular} & \begin{tabular}[c]{@{}c@{}}${\text{MSE}}_\text{V}$\\ \cite{ref17}\end{tabular} & \begin{tabular}[c]{@{}c@{}}${\text{PSNR}}_\text{Y}$\\ \cite{ref17}\end{tabular} & \begin{tabular}[c]{@{}c@{}}$\ADD{{\text{PSNR}}_\text{U}}$\\ \cite{ref17}\end{tabular} & \multicolumn{1}{c|}{\begin{tabular}[c]{@{}c@{}}$\ADD{{\text{PSNR}}_\text{V}}$\\ \cite{ref17}\end{tabular}} & \multirow{-2}{*}{\bf{RBFIM}}   \\ \hline
\multicolumn{1}{c|}{}                                                                            &                         & PLCC                       & {\color[HTML]{3531FF} 0.880}                                & 0.872                                                           & 0.629                                                   & {\color[HTML]{FF0000} 0.917}                                   & \multicolumn{1}{c|}{0.696}                                                          & 0.617                                                       & 0.445                                                        & 0.572                                                       & 0.457                                                        & 0.472                                                    & 0.436                                                    & 0.346                                                    & 0.715                                                     & 0.563                                                     & \multicolumn{1}{c|}{0.690}                                                     & 0.788                        \\
\multicolumn{1}{c|}{}                                                                            &                         & SROCC                      & 0.882                                                       & 0.874                                                           & {\color[HTML]{3531FF} 0.934}                            & {\color[HTML]{FF0000} 0.952}                                   & \multicolumn{1}{c|}{0.819}                                                          & 0.921                                                       & 0.539                                                        & 0.930                                                       & 0.544                                                        & 0.723                                                    & 0.559                                                    & 0.738                                                    & 0.723                                                     & 0.559                                                     & \multicolumn{1}{c|}{0.738}                                                     & 0.802                        \\
\multicolumn{1}{c|}{}                                                                            &                         & KROCC                      & 0.703                                                       & 0.689                                                           & 0.774                                                   & {\color[HTML]{FF0000} 0.811}                                   & \multicolumn{1}{c|}{0.652}                                                          & 0.753                                                       & 0.415                                                        & {\color[HTML]{3531FF} 0.782}                                & 0.419                                                        & 0.547                                                    & 0.405                                                    & 0.544                                                    & 0.547                                                     & 0.405                                                     & \multicolumn{1}{c|}{0.543}                                                     & 0.616                        \\
\multicolumn{1}{c|}{}                                                                            & \multirow{-4}{*}{G-PCC} & RMSE                       & 0.276                                                       & 0.331                                                           & {\color[HTML]{3531FF} 0.205}                            & {\color[HTML]{FF0000} 0.180}                                   & \multicolumn{1}{c|}{0.343}                                                          & 0.275                                                       & 0.671                                                        & 0.230                                                       & 0.647                                                        & 0.564                                                    & 0.587                                                    & 0.368                                                    & 0.567                                                     & 0.576                                                     & \multicolumn{1}{c|}{0.448}                                                     & 0.368                        \\ \cline{2-19} 
\multicolumn{1}{c|}{}                                                                            &                         & PLCC                       & {\color[HTML]{FF0000} 0.808}                                & {\color[HTML]{3531FF} 0.725}                                    & 0.706                                                   & 0.233                                                          & \multicolumn{1}{c|}{0.271}                                                          & 0.414                                                       & 0.196                                                        & 0.612                                                       & 0.254                                                        & 0.139                                                    & 0.049                                                    & 0.355                                                    & 0.303                                                     & 0.085                                                     & \multicolumn{1}{c|}{0.326}                                                     & 0.469                        \\
\multicolumn{1}{c|}{}                                                                            &                         & SROCC                      & {\color[HTML]{FF0000} 0.819}                                & 0.735                                                           & {\color[HTML]{3531FF} 0.755}                            & 0.116                                                          & \multicolumn{1}{c|}{0.387}                                                          & 0.420                                                       & 0.246                                                        & 0.691                                                       & 0.327                                                        & 0.333                                                    & 0.114                                                    & 0.300                                                    & 0.333                                                     & 0.114                                                     & \multicolumn{1}{c|}{0.302}                                                     & 0.475                        \\
\multicolumn{1}{c|}{}                                                                            &                         & KROCC                      & {\color[HTML]{FF0000} 0.616}                                & 0.526                                                           & {\color[HTML]{3531FF} 0.549}                            & 0.067                                                          & \multicolumn{1}{c|}{0.281}                                                          & 0.273                                                       & 0.186                                                        & 0.500                                                       & 0.232                                                        & 0.235                                                    & 0.075                                                    & 0.210                                                    & 0.235                                                     & 0.075                                                     & \multicolumn{1}{c|}{0.211}                                                     & 0.320                        \\
\multicolumn{1}{c|}{\multirow{-8}{*}{\begin{tabular}[c]{@{}c@{}}M-PCCD\\ \cite{ref51}\end{tabular}}} & \multirow{-4}{*}{V-PCC} & RMSE                       & {\color[HTML]{3531FF} 0.306}                                & {\color[HTML]{FF0000} 0.305}                                    & 0.441                                                   & 0.556                                                          & \multicolumn{1}{c|}{0.557}                                                          & 0.529                                                       & 0.556                                                        & 0.496                                                       & 0.556                                                        & 0.556                                                    & 0.556                                                    & 0.469                                                    & 0.519                                                     & 0.556                                                     & \multicolumn{1}{c|}{0.509}                                                     & 0.588                        \\ \hline
\multicolumn{1}{c|}{}                                                                            &                         & PLCC                       & 0.707                                                       & 0.669                                                           & 0.763                                                   & {\color[HTML]{FF0000} 0.978}                                   & \multicolumn{1}{c|}{0.573}                                                          & 0.772                                                       & 0.695                                                        & 0.734                                                       & 0.710                                                        & 0.722                                                    & 0.494                                                    & 0.518                                                    & 0.917                                                     & 0.801                                                     & \multicolumn{1}{c|}{0.739}                                                     & {\color[HTML]{3531FF} 0.963} \\
\multicolumn{1}{c|}{}                                                                            &                         & SROCC                      & 0.635                                                       & 0.616                                                           & {\color[HTML]{FF0000} 0.972}                            & 0.894                                                          & \multicolumn{1}{c|}{0.646}                                                          & 0.886                                                       & 0.705                                                        & 0.911                                                       & 0.735                                                        & 0.908                                                    & 0.767                                                    & 0.689                                                    & 0.908                                                     & 0.767                                                     & \multicolumn{1}{c|}{0.689}                                                     & {\color[HTML]{3531FF} 0.941} \\
\multicolumn{1}{c|}{}                                                                            &                         & KROCC                      & 0.444                                                       & 0.449                                                           & {\color[HTML]{FF0000} 0.865}                            & 0.740                                                          & \multicolumn{1}{c|}{0.476}                                                          & 0.712                                                       & 0.550                                                        & 0.754                                                       & 0.578                                                        & 0.749                                                    & 0.573                                                    & 0.509                                                    & 0.749                                                     & 0.573                                                     & \multicolumn{1}{c|}{0.509}                                                     & {\color[HTML]{3531FF} 0.814} \\
\multicolumn{1}{c|}{}                                                                            & \multirow{-4}{*}{G-PCC} & RMSE                       & 0.647                                                       & 0.671                                                           & {\color[HTML]{FF0000} 0.180}                            & 0.331                                                          & \multicolumn{1}{c|}{0.587}                                                          & 0.343                                                       & 0.564                                                        & 0.230                                                       & 0.448                                                        & 0.276                                                    & 0.368                                                    & 0.576                                                    & 0.275                                                     & 0.368                                                     & \multicolumn{1}{c|}{0.567}                                                     & {\color[HTML]{3531FF} 0.205} \\ \cline{2-19} 
\multicolumn{1}{c|}{}                                                                            &                         & PLCC                       & 0.796                                                       & 0.798                                                           & {\color[HTML]{3531FF} 0.940}                            & 0.799                                                          & \multicolumn{1}{c|}{0.337}                                                          & 0.896                                                       & 0.466                                                        & 0.920                                                       & 0.517                                                        & 0.704                                                    & 0.408                                                    & 0.441                                                    & 0.894                                                     & 0.559                                                     & \multicolumn{1}{c|}{0.488}                                                     & {\color[HTML]{FF0000} 0.956} \\
\multicolumn{1}{c|}{}                                                                            &                         & SROCC                      & 0.792                                                       & 0.770                                                           & {\color[HTML]{FF0000} 0.958}                            & 0.754                                                          & \multicolumn{1}{c|}{0.444}                                                          & 0.872                                                       & 0.554                                                        & 0.907                                                       & 0.580                                                        & 0.881                                                    & 0.562                                                    & 0.446                                                    & 0.881                                                     & 0.562                                                     & \multicolumn{1}{c|}{0.448}                                                     & {\color[HTML]{3531FF} 0.952} \\
\multicolumn{1}{c|}{}                                                                            &                         & KROCC                      & 0.618                                                       & 0.567                                                           & {\color[HTML]{FF0000} 0.830}                            & 0.595                                                          & \multicolumn{1}{c|}{0.318}                                                          & 0.669                                                       & 0.438                                                        & 0.720                                                       & 0.461                                                        & 0.692                                                    & 0.388                                                    & 0.321                                                    & 0.692                                                     & 0.388                                                     & \multicolumn{1}{c|}{0.323}                                                     & {\color[HTML]{3531FF} 0.826} \\
\multicolumn{1}{c|}{\multirow{-8}{*}{\begin{tabular}[c]{@{}c@{}}ICIP\\ \cite{ref49}\end{tabular}}}   & \multirow{-4}{*}{V-PCC} & RMSE                       & 0.307                                                       & 0.307                                                           & 0.239                                                   & 0.307                                                          & \multicolumn{1}{c|}{0.470}                                                          & {\color[HTML]{3531FF} 0.236}                                & 0.586                                                        & 0.239                                                       & 0.527                                                        & 0.442                                                    & 0.526                                                    & 0.526                                                    & {\color[HTML]{3531FF} 0.236}                              & 0.507                                                     & \multicolumn{1}{c|}{0.580}                                                     & {\color[HTML]{FF0000} 0.224} \\ \hline
\multicolumn{1}{c|}{}                                                                            &                         & PLCC                       & 0.723                                                       & 0.767                                                           & 0.411                                                   & 0.512                                                          & \multicolumn{1}{c|}{0.528}                                                          & 0.501                                                       & 0.188                                                        & 0.454                                                       & 0.291                                                        & 0.784                                                    & 0.448                                                    & 0.266                                                    & {\color[HTML]{FF0000} 0.819}                              & 0.617                                                     & \multicolumn{1}{c|}{0.503}                                                     & {\color[HTML]{3531FF} 0.815} \\
\multicolumn{1}{c|}{}                                                                            &                         & SROCC                      & 0.812                                                       & {\color[HTML]{3531FF} 0.853}                                    & 0.443                                                   & 0.418                                                          & \multicolumn{1}{c|}{0.598}                                                          & 0.789                                                       & 0.195                                                        & 0.806                                                       & 0.245                                                        & 0.847                                                    & 0.672                                                    & 0.507                                                    & 0.847                                                     & 0.672                                                     & \multicolumn{1}{c|}{0.507}                                                     & {\color[HTML]{FF0000} 0.886} \\
\multicolumn{1}{c|}{}                                                                            &                         & KROCC                      & 0.608                                                       & {\color[HTML]{3531FF} 0.700}                                    & 0.320                                                   & 0.294                                                          & \multicolumn{1}{c|}{0.451}                                                          & 0.634                                                       & 0.085                                                        & 0.660                                                       & 0.111                                                        & 0.673                                                    & 0.477                                                    & 0.386                                                    & 0.673                                                     & 0.477                                                     & \multicolumn{1}{c|}{0.386}                                                     & {\color[HTML]{FF0000} 0.752} \\
\multicolumn{1}{c|}{}                                                                            & \multirow{-4}{*}{G-PCC} & RMSE                       & 0.292                                                       & {\color[HTML]{3531FF} 0.289}                                    & 0.625                                                   & 0.577                                                          & \multicolumn{1}{c|}{0.326}                                                          & 0.662                                                       & 0.605                                                        & 0.649                                                       & 0.628                                                        & 0.335                                                    & 0.420                                                    & 0.417                                                    & 0.333                                                     & 0.470                                                     & \multicolumn{1}{c|}{0.401}                                                     & {\color[HTML]{FF0000} 0.265} \\ \cline{2-19} 
\multicolumn{1}{c|}{}                                                                            &                         & PLCC                       & 0.336                                                       & 0.338                                                           & 0.291                                                   & 0.255                                                          & \multicolumn{1}{c|}{0.428}                                                          & 0.017                                                       & 0.330                                                        & 0.316                                                       & 0.255                                                        & 0.575                                                    & {\color[HTML]{3531FF} 0.629}                             & {\color[HTML]{FF0000} 0.631}                             & 0.406                                                     & 0.476                                                     & \multicolumn{1}{c|}{0.456}                                                     & 0.451                        \\
\multicolumn{1}{c|}{}                                                                            &                         & SROCC                      & 0.511                                                       & 0.497                                                           & 0.337                                                   & 0.162                                                          & \multicolumn{1}{c|}{0.512}                                                          & 0.131                                                       & 0.506                                                        & 0.087                                                       & 0.217                                                        & {\color[HTML]{FF0000} 0.703}                             & 0.566                                                    & 0.585                                                    & {\color[HTML]{FF0000} 0.703}                              & 0.566                                                     & \multicolumn{1}{c|}{0.585}                                                     & {\color[HTML]{3531FF} 0.632}                        \\
\multicolumn{1}{c|}{}                                                                            &                         & KROCC                      & 0.415                                                       & 0.397                                                           & 0.197                                                   & 0.116                                                          & \multicolumn{1}{c|}{0.402}                                                          & 0.075                                                       & 0.361                                                        & 0.007                                                       & 0.129                                                        & {\color[HTML]{FF0000} 0.565}                             & 0.415                                                    & 0.443                                                    & {\color[HTML]{FF0000} 0.565}                              & 0.415                                                     & \multicolumn{1}{c|}{0.443}                                                     & {\color[HTML]{3531FF} 0.483}                        \\
\multicolumn{1}{c|}{\multirow{-8}{*}{\begin{tabular}[c]{@{}c@{}}IRPC\\ \cite{ref50}\end{tabular}}}   & \multirow{-4}{*}{V-PCC} & RMSE                       & {\color[HTML]{FF0000} 0.470}                                & {\color[HTML]{FF0000} 0.470}                                    & 0.555                                                   & 0.556                                                          & \multicolumn{1}{c|}{0.529}                                                          & 0.556                                                       & {\color[HTML]{3531FF} 0.473}                                                        & 0.518                                                       & 0.556                                                        & 0.517                                                    & 0.492                                                    & 0.497                                                    & 0.526                                                     & 0.588                                                     & \multicolumn{1}{c|}{0.580}                                                     & 0.570                        \\ \hline
\multicolumn{1}{c|}{}                                                                            &                         & PLCC                       & 0.645                                                       & 0.555                                                           & 0.722                                                   & 0.757                                                          & \multicolumn{1}{c|}{{\color[HTML]{3531FF} 0.892}}                                   & 0.341                                                       & 0.501                                                        & 0.358                                                       & 0.530                                                        & 0.713                                                    & 0.343                                                    & 0.293                                                    & 0.824                                                     & 0.568                                                     & \multicolumn{1}{c|}{0.542}                                                     & {\color[HTML]{FF0000} 0.907} \\
\multicolumn{1}{c|}{}                                                                            &                         & SROCC                      & 0.628                                                       & 0.535                                                           & 0.883                                                   & 0.712                                                          & \multicolumn{1}{c|}{{\color[HTML]{FF0000} 0.934}}                                   & 0.660                                                       & 0.537                                                        & 0.682                                                       & 0.551                                                        & 0.845                                                    & 0.620                                                    & 0.646                                                    & 0.845                                                     & 0.620                                                     & \multicolumn{1}{c|}{0.646}                                                     & {\color[HTML]{3531FF} 0.924} \\
\multicolumn{1}{c|}{}                                                                            &                         & KROCC                      & 0.509                                                       & 0.411                                                           & 0.723                                                   & 0.518                                                          & \multicolumn{1}{c|}{{\color[HTML]{FF0000} 0.770}}                                   & 0.468                                                       & 0.376                                                        & 0.489                                                       & 0.386                                                        & 0.652                                                    & 0.439                                                    & 0.475                                                    & 0.652                                                     & 0.439                                                     & \multicolumn{1}{c|}{0.474}                                                     & {\color[HTML]{3531FF} 0.759} \\
\multicolumn{1}{c|}{\multirow{-4}{*}{\begin{tabular}[c]{@{}c@{}}NWPU\\ \cite{ref48}\end{tabular}}}   & \multirow{-4}{*}{G-PCC} & RMSE                       & 0.566                                                       & 0.608                                                           & 0.241                                                   & 0.295                                                          & \multicolumn{1}{c|}{{\color[HTML]{FF0000} 0.170}}                                   & 0.305                                                       & 0.608                                                        & 0.297                                                       & 0.593                                                        & 0.282                                                    & 0.576                                                    & 0.339                                                    & 0.260                                                     & 0.585                                                     & \multicolumn{1}{c|}{0.307}                                                     & {\color[HTML]{3531FF} 0.196} \\ \hline
\multicolumn{1}{c|}{}                                                                            &                         & PLCC                       & 0.827                                                       & 0.868                                                           & {\color[HTML]{3531FF} 0.871}                            & --                                                             & \multicolumn{1}{c|}{0.768}                                                          & --                                                          & --                                                           & --                                                          & --                                                           & 0.741                                                    & 0.662                                                    & 0.590                                                    & 0.796                                                     & 0.724                                                     & \multicolumn{1}{c|}{0.684}                                                     & {\color[HTML]{FF0000} 0.873} \\
\multicolumn{1}{c|}{}                                                                            &                         & SROCC                      & 0.830                                                       & 0.865                                                           & {\color[HTML]{3531FF} 0.894}                            & --                                                             & \multicolumn{1}{c|}{0.795}                                                          & --                                                          & --                                                           & --                                                          & --                                                           & 0.810                                                    & 0.755                                                    & 0.713                                                    & 0.810                                                     & 0.755                                                     & \multicolumn{1}{c|}{0.713}                                                     & {\color[HTML]{FF0000} 0.898} \\
\multicolumn{1}{c|}{}                                                                            &                         & KROCC                      & 0.632                                                       & 0.671                                                           & {\color[HTML]{3531FF} 0.704}                            & --                                                             & \multicolumn{1}{c|}{0.595}                                                          & --                                                          & --                                                           & --                                                          & --                                                           & 0.603                                                    & 0.573                                                    & 0.530                                                    & 0.603                                                     & 0.573                                                     & \multicolumn{1}{c|}{0.530}                                                     & {\color[HTML]{FF0000} 0.708} \\
\multicolumn{1}{c|}{}                                                                            & \multirow{-4}{*}{G-PCC} & RMSE                       & 0.603                                                       & 0.594                                                           & {\color[HTML]{3531FF} 0.143}                                                   & --                                                             & \multicolumn{1}{c|}{0.164}                                                          & --                                                          & --                                                           & --                                                          & --                                                           & 0.209                                                    & 0.227                                                    & 0.206                                                    & 0.226                                                     & 0.636                                                     & \multicolumn{1}{c|}{0.617}                                                     & {\color[HTML]{FF0000} 0.131} \\ \cline{2-19} 
\multicolumn{1}{c|}{}                                                                            &                         & PLCC                       & 0.636                                                       & {\color[HTML]{FF0000} 0.717}                                    & 0.561                                                   & 0.084                                                          & \multicolumn{1}{c|}{0.316}                                                          & 0.684                                                       & 0.167                                                        & {\color[HTML]{3531FF} 0.705}                                & 0.211                                                        & 0.296                                                    & 0.249                                                    & 0.213                                                    & 0.349                                                     & 0.233                                                     & \multicolumn{1}{c|}{0.230}                                                     & 0.506                        \\
\multicolumn{1}{c|}{}                                                                            &                         & SROCC                      & 0.609                                                       & 0.692                                                           & 0.553                                                   & 0.098                                                          & \multicolumn{1}{c|}{0.326}                                                          & {\color[HTML]{3531FF} 0.698}                                & 0.160                                                        & {\color[HTML]{FF0000} 0.707}                                & 0.200                                                        & 0.342                                                    & 0.233                                                    & 0.237                                                    & 0.342                                                     & 0.233                                                     & \multicolumn{1}{c|}{0.237}                                                     & 0.500                        \\
\multicolumn{1}{c|}{}                                                                            &                         & KROCC                      & 0.434                                                       & {\color[HTML]{FF0000} 0.509}                                    & 0.402                                                   & 0.065                                                          & \multicolumn{1}{c|}{0.224}                                                          & 0.498                                                       & 0.111                                                        & {\color[HTML]{3531FF} 0.502}                                & 0.136                                                        & 0.237                                                    & 0.157                                                    & 0.158                                                    & 0.237                                                     & 0.157                                                     & \multicolumn{1}{c|}{0.158}                                                     & 0.356                        \\
\multicolumn{1}{c|}{\multirow{-8}{*}{\begin{tabular}[c]{@{}c@{}}WPC\\ \cite{ref28}\end{tabular}}}    & \multirow{-4}{*}{V-PCC} & RMSE                       & 0.497                                                       & {\color[HTML]{FF0000} 0.310}                                    & 0.511                                                   & 0.556                                                          & \multicolumn{1}{c|}{0.518}                                                          & {\color[HTML]{3531FF} 0.414}                                & 0.556                                                        & 0.441                                                       & 0.556                                                        & 0.519                                                    & 0.556                                                    & 0.556                                                    & 0.476                                                     & 0.556                                                     & \multicolumn{1}{c|}{0.556}                                                     & 0.536                        \\ \hline
                                                                                                 &                         &                            &                                                             &                                                                 &                                                         &                                                                &                                                                                     &                                                             &                                                              &                                                             &                                                              &                                                          &                                                          &                                                          &                                                           &                                                           &                                                                                &                              \\ \hline
\multicolumn{2}{c|}{}                                                                                                      & PLCC                       & 0.695                                                       & 0.611                                                           & 0.681                                                   & 0.673                                                          & \multicolumn{1}{c|}{0.666}                                                          & 0.273                                                       & 0.484                                                        & 0.292                                                       & 0.505                                                        & 0.539                                                    & 0.319                                                    & 0.318                                                    & {\color[HTML]{3531FF} 0.784}                              & 0.462                                                     & \multicolumn{1}{c|}{0.467}                                                     & {\color[HTML]{FF0000} 0.870} \\
\multicolumn{2}{c|}{}                                                                                                      & SROCC                      & 0.690                                                       & 0.603                                                           & {\color[HTML]{3531FF} 0.877}                            & 0.555                                                          & \multicolumn{1}{c|}{0.853}                                                          & 0.496                                                       & 0.503                                                        & 0.500                                                       & 0.514                                                        & 0.809                                                    & 0.492                                                    & 0.506                                                    & 0.809                                                     & 0.491                                                     & \multicolumn{1}{c|}{0.506}                                                     & {\color[HTML]{FF0000} 0.888} \\
\multicolumn{2}{c|}{}                                                                                                      & KROCC                      & 0.542                                                       & 0.454                                                           & {\color[HTML]{3531FF} 0.692}                            & 0.392                                                          & \multicolumn{1}{c|}{0.660}                                                          & 0.339                                                       & 0.351                                                        & 0.344                                                       & 0.359                                                        & 0.611                                                    & 0.336                                                    & 0.357                                                    & 0.611                                                     & 0.336                                                     & \multicolumn{1}{c|}{0.357}                                                     & {\color[HTML]{FF0000} 0.702} \\
\multicolumn{2}{c|}{\multirow{-4}{*}{ALL G-PCC}}                                                                           & RMSE                       & 0.290                                                       & 0.445                                                           & {\color[HTML]{3531FF} 0.221}                            & 0.585                                                          & \multicolumn{1}{c|}{0.261}                                                          & 0.627                                                       & 0.614                                                        & 0.616                                                       & 0.592                                                        & 0.285                                                    & 0.633                                                    & 0.601                                                    & 0.276                                                     & 0.632                                                     & \multicolumn{1}{c|}{0.597}                                                     & {\color[HTML]{FF0000} 0.200} \\ \hline
\multicolumn{2}{c|}{}                                                                                                      & PLCC                       & {\color[HTML]{3531FF} 0.450}                                & 0.406                                                           & 0.249                                                   & 0.209                                                          & \multicolumn{1}{c|}{0.113}                                                          & 0.298                                                       & 0.097                                                        & 0.273                                                       & 0.031                                                        & 0.224                                                    & 0.072                                                    & 0.050                                                    & 0.188                                                     & 0.235                                                     & \multicolumn{1}{c|}{0.268}                                                     & {\color[HTML]{FF0000} 0.510} \\
\multicolumn{2}{c|}{}                                                                                                      & SROCC                      & 0.483                                                       & {\color[HTML]{FF0000} 0.518}                                    & 0.480                                                   & 0.089                                                          & \multicolumn{1}{c|}{0.236}                                                          & 0.410                                                       & 0.046                                                        & 0.425                                                       & 0.089                                                        & 0.293                                                    & 0.221                                                    & 0.256                                                    & 0.293                                                     & 0.221                                                     & \multicolumn{1}{c|}{0.256}                                                     & {\color[HTML]{3531FF} 0.511} \\
\multicolumn{2}{c|}{}                                                                                                      & KROCC                      & {\color[HTML]{3531FF} 0.359}                                & 0.392                                                           & 0.357                                                   & 0.060                                                          & \multicolumn{1}{c|}{0.169}                                                          & 0.305                                                       & 0.034                                                        & 0.315                                                       & 0.064                                                        & 0.212                                                    & 0.152                                                    & 0.175                                                    & 0.212                                                     & 0.152                                                     & \multicolumn{1}{c|}{0.175}                                                     & {\color[HTML]{FF0000} 0.368} \\
\multicolumn{2}{c|}{\multirow{-4}{*}{ALL V-PCC}}                                                                           & RMSE                       & {\color[HTML]{3531FF} 0.526}                                & {\color[HTML]{3531FF} 0.526}                                    & 0.556                                                   & 0.556                                                          & \multicolumn{1}{c|}{0.556}                                                          & 0.529                                                       & 0.556                                                        & 0.557                                                       & 0.556                                                        & 0.556                                                    & 0.556                                                    & 0.556                                                    & 0.556                                                     & 0.556                                                     & \multicolumn{1}{c|}{0.556}                                                     & {\color[HTML]{FF0000} 0.516} \\ \hline
\multicolumn{2}{c|}{}                                                                                                      & PLCC                       & {\color[HTML]{3531FF} 0.640}                                & 0.563                                                           & 0.053                                                   & 0.476                                                          & \multicolumn{1}{c|}{0.302}                                                          & 0.132                                                       & 0.298                                                        & 0.121                                                       & 0.348                                                        & 0.014                                                    & 0.137                                                    & 0.264                                                    & 0.578                                                     & 0.409                                                     & \multicolumn{1}{c|}{0.427}                                                     & {\color[HTML]{FF0000} 0.764} \\
\multicolumn{2}{c|}{}                                                                                                      & SROCC                      & 0.651                                                       & 0.583                                                           & {\color[HTML]{FF0000} 0.786}                            & 0.429                                                          & \multicolumn{1}{c|}{0.673}                                                          & 0.511                                                       & 0.319                                                        & 0.524                                                       & 0.363                                                        & 0.663                                                    & 0.409                                                    & 0.419                                                    & 0.663                                                     & 0.409                                                     & \multicolumn{1}{c|}{0.419}                                                     & {\color[HTML]{3531FF} 0.765} \\
\multicolumn{2}{c|}{}                                                                                                      & KROCC                      & 0.505                                                       & 0.444                                                           & {\color[HTML]{FF0000} 0.612}                            & 0.299                                                          & \multicolumn{1}{c|}{0.501}                                                          & 0.366                                                       & 0.222                                                        & 0.376                                                       & 0.256                                                        & 0.485                                                    & 0.274                                                    & 0.287                                                    & 0.485                                                     & 0.274                                                     & \multicolumn{1}{c|}{0.287}                                                     & {\color[HTML]{3531FF} 0.575} \\
\multicolumn{2}{c|}{\multirow{-4}{*}{ALL}}                                                                                 & RMSE                       & 0.497                                                       & 0.516                                                           & {\color[HTML]{3531FF} 0.316}                            & 0.538                                                          & \multicolumn{1}{c|}{0.490}                                                          & 0.533                                                       & 0.520                                                        & 0.464                                                       & 0.485                                                        & 0.470                                                    & 0.533                                                    & 0.538                                                    & 0.470                                                     & 0.533                                                     & \multicolumn{1}{c|}{0.538}                                                     & {\color[HTML]{FF0000} 0.309} \\ \hline
\end{tabular}
\end{adjustbox}
\label{table2}
\end{table*}
\begin{figure*}[t]
  \centering
  \subfloat[PLCC]{\label{fig:a}\includegraphics[width=3.1in]{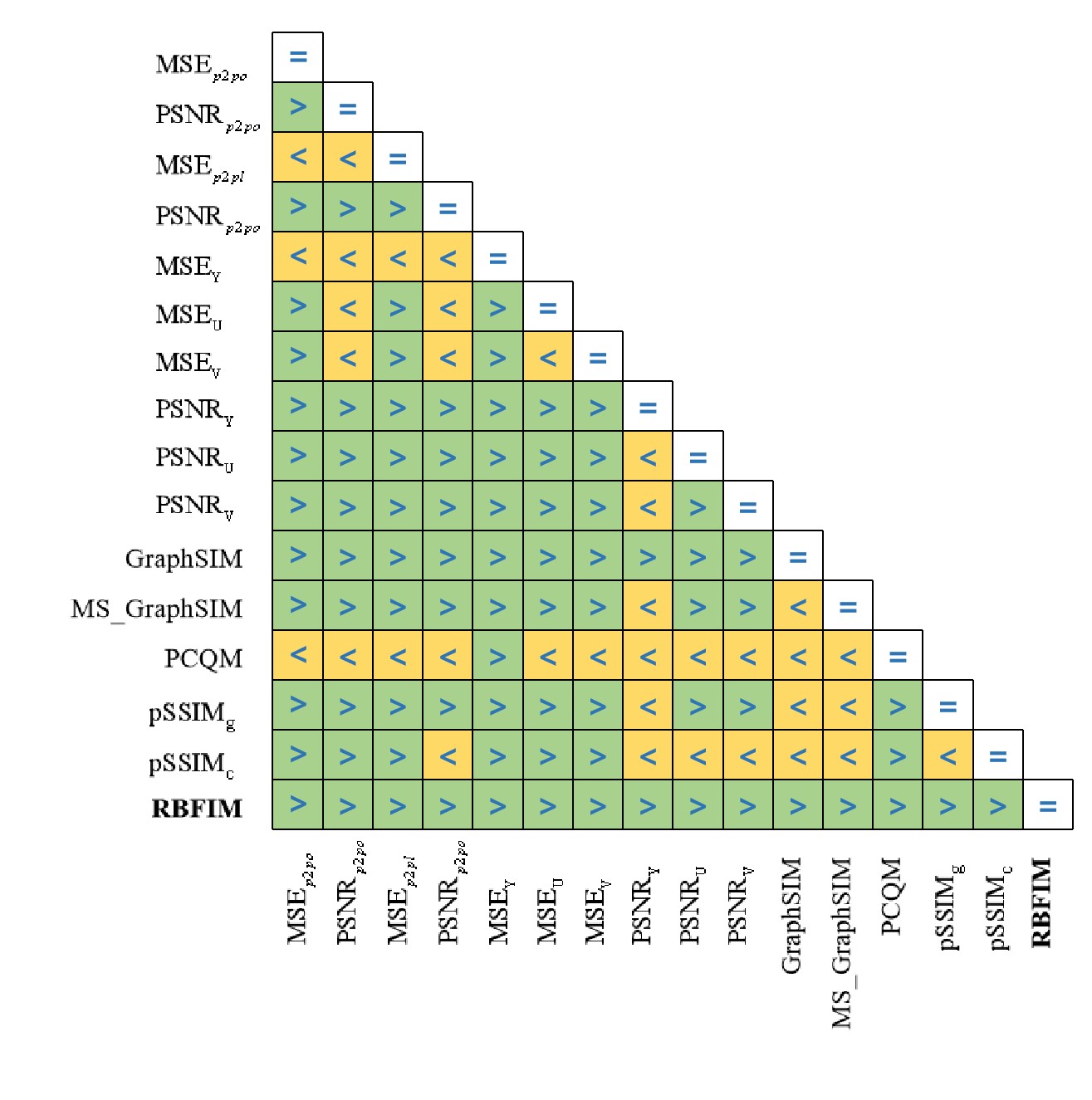}}\quad
  \subfloat[SROCC]{\label{fig:i}\includegraphics[width=3.1in]{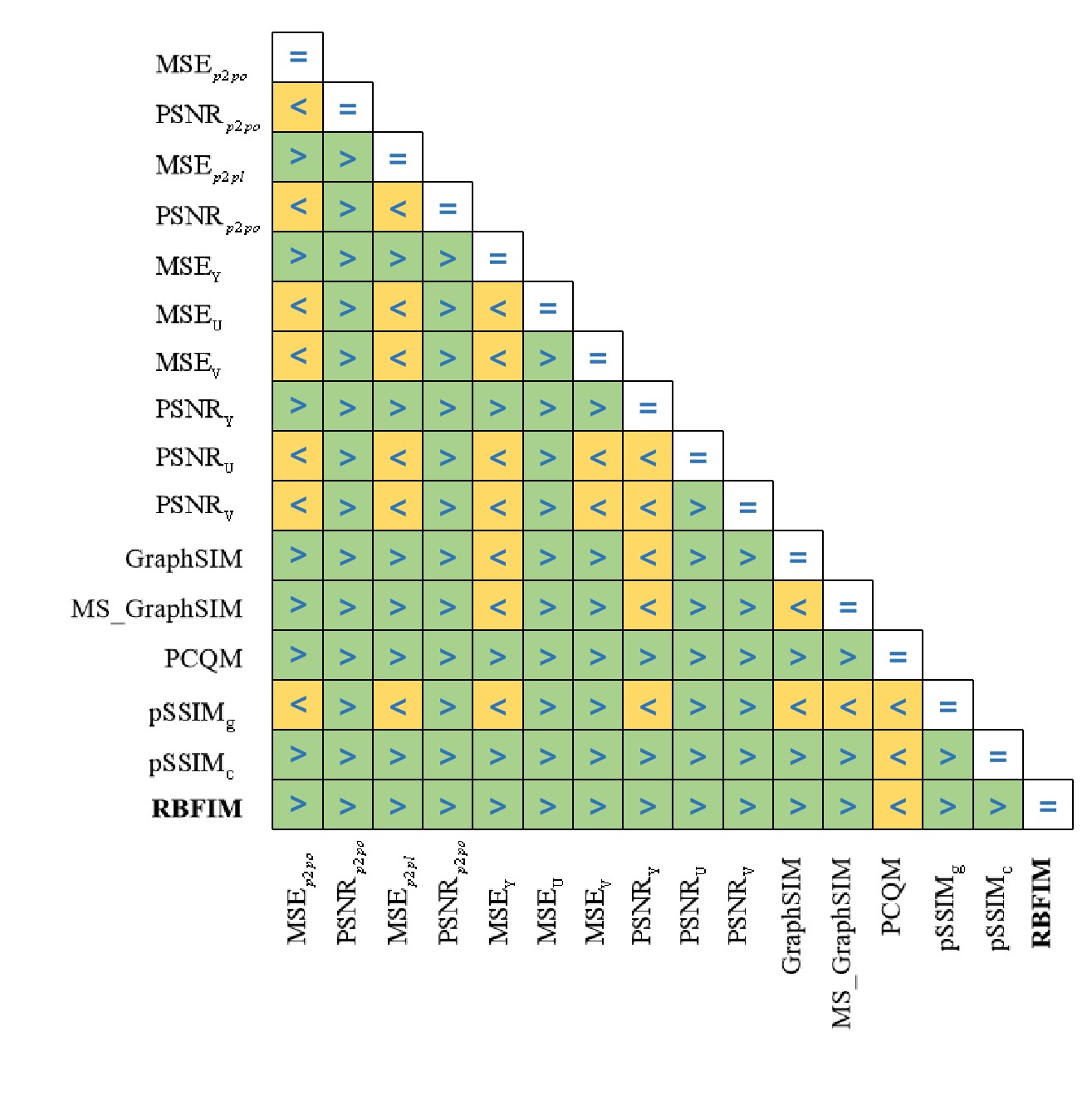}}\quad
  \subfloat[KROCC]{\label{fig:e}\includegraphics[width=3.1in]{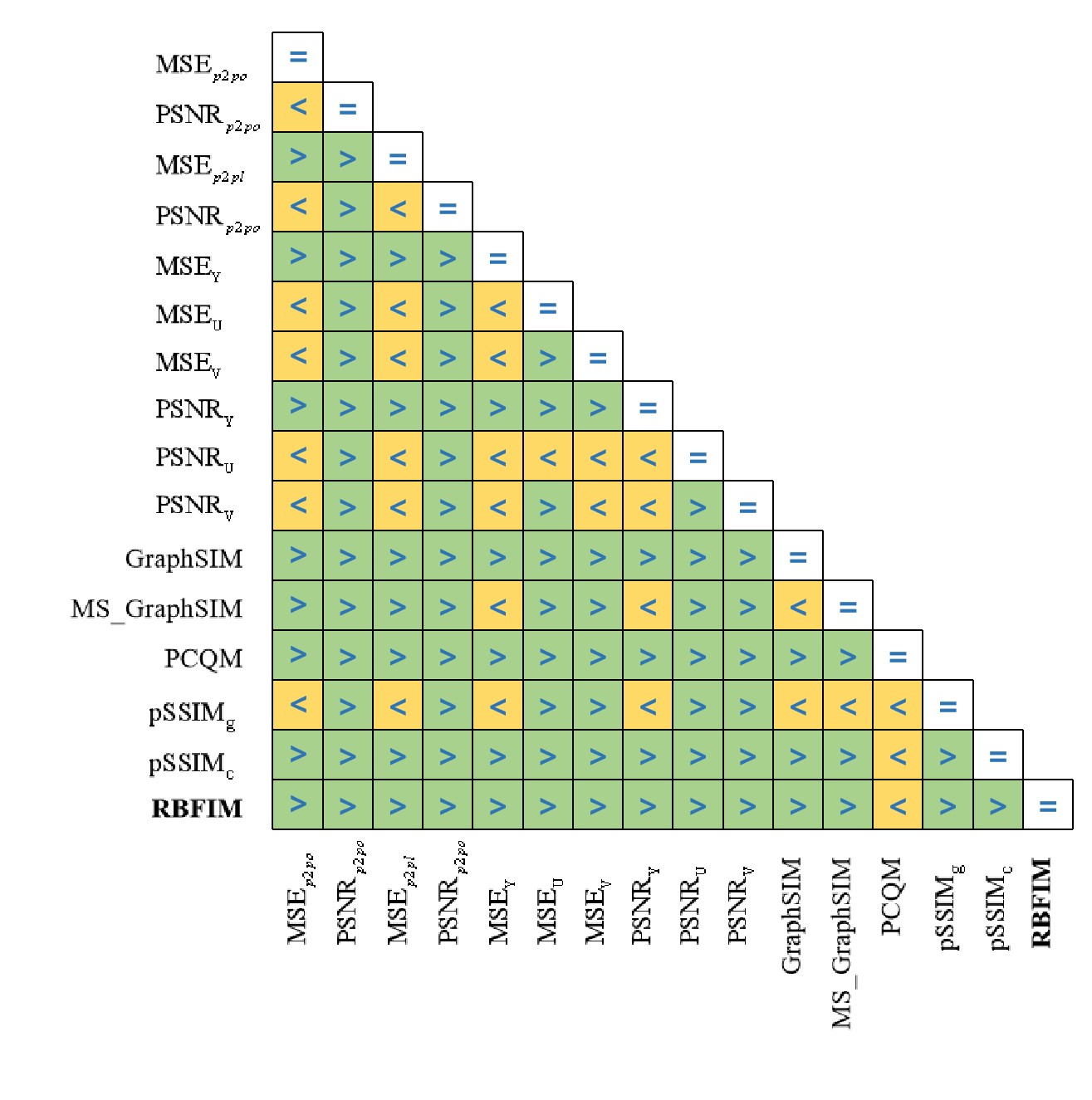}}\quad
  \subfloat[\ADD{RMSE}]{\label{fig:i}\includegraphics[width=3.1in]{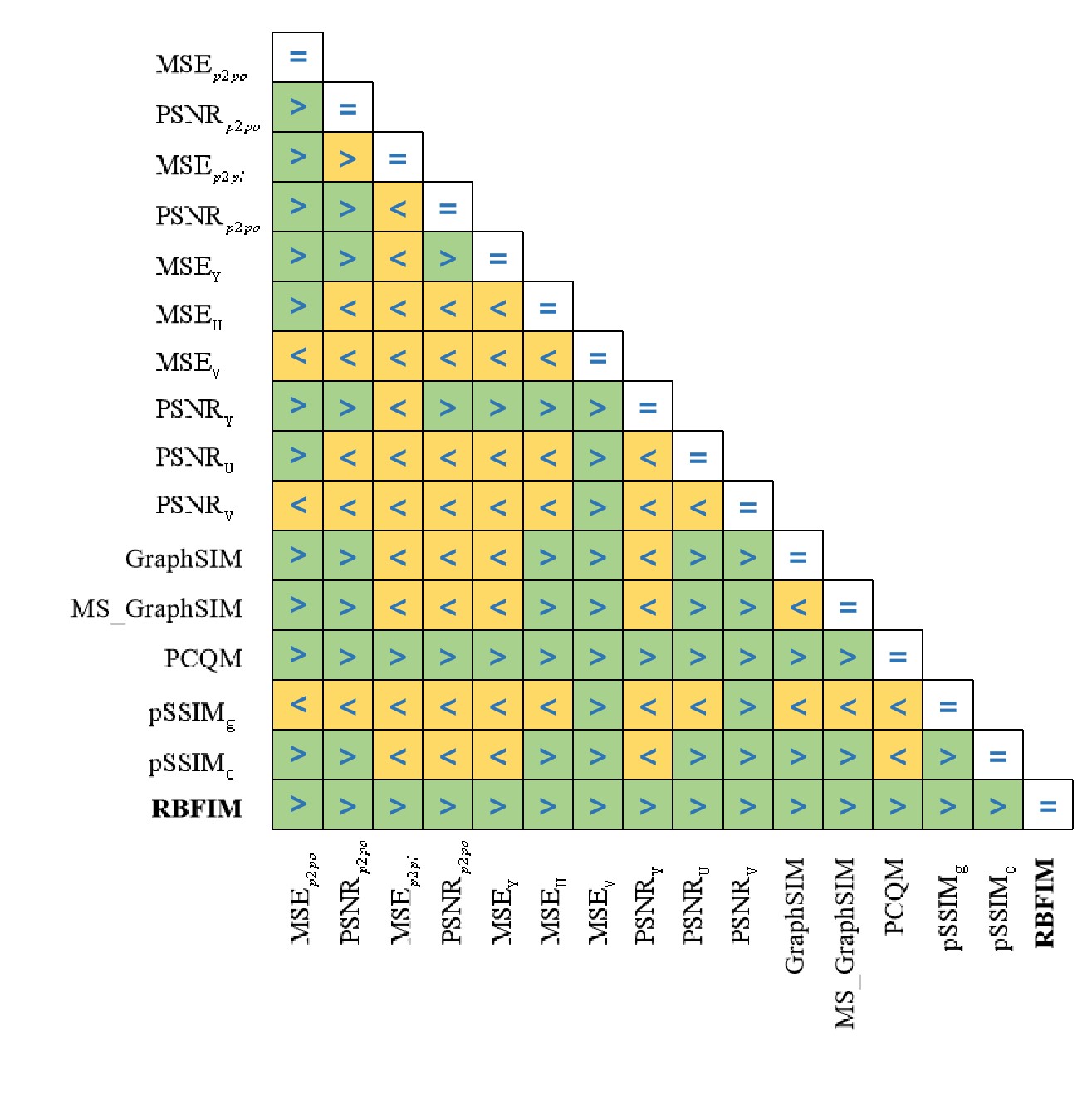}}\\
  \caption{Performance comparison between different metrics. The "\textgreater" symbol (highlighted in green) indicates that the metric in the row is superior to the one in the column, and the "\textless" symbol (highlighted in yellow) indicates that the metric in the row is inferior to the one in the column.}
  \label{fig3}
  \end{figure*}
\subsection{Evaluation Metrics}
To ensure alignment between the subjective ratings and objective predictions of different metrics, we standardize the objective predictions to a consistent dynamic range based on guidance from the video quality expert group (VQEG) \cite{ref47}. Subsequently, we utilize Pearson's linear correlation coefficient (PLCC), Spearman rank order correlation coefficient (SROCC), Kendall rank order correlation coefficient (KROCC), and root mean square error (RMSE) to evaluate the performance of various metrics, representing their linearity, monotonicity, and accuracy, respectively. Higher PLCC, SROCC and KROCC values indicate superior metric performance, while lower RMSE values suggest better accuracy. To normalize the scores of objective quality assessment metrics onto a uniform scale, we apply the logistic regression method recommended by VQEG.\par
\subsection{Datasets and Methods under Comparison}
To verify the performance of RBFIM in terms of compression distortion, we use the G-PCC and V-PCC distortion from five datasets for validation. The datasets include: NWPU-PCQD \cite{ref48}, WPC \cite{ref28}, ICIP2020 \cite{ref49}, IRPC \cite{ref50} and M-PCCD\cite{ref51}. Additionally, these five datasets are integrated into a comprehensive dataset to assess the metrics' stability and reliability. The proposed RBFIM is compared with 15 PCQA metrics. These metrics include single-feature metrics: ${\text{MSE}}_{p2po}$\cite{ref17}, ${\text{PSNR}}_{p2po}$\cite{ref17}, ${\text{MSE}}_{p2pl}$\cite{ref17}, ${\text{PSNR}}_{p2pl}$\cite{ref17}, ${\text{MSE}}_\text{Y}$\cite{ref17}, ${\text{MSE}}_\text{U}$\cite{ref17}, ${\text{MSE}}_\text{V}$\cite{ref17}, ${\text{PSNR}}_\text{Y}$\cite{ref17}, ${\text{PSNR}}_\text{U}$\cite{ref17}, and ${\text{PSNR}}_\text{V}$\cite{ref17} and multi-feature metrics: PCQM\cite{ref31}, GraphSIM\cite{ref33}, MS-GraphSIM\cite{ref34}, ${\text{pSSIM}}_\text{c}$\cite{ref35}, and ${\text{pSSIM}}_\text{g}$\cite{ref35}.\par
\subsection{Parameter Settings}
\begin{itemize}
\item{During the process of content adaptive partition, we chose ${T_{\max }} = 40$ and ${T_{\min }} = 20$ as the respective maximum and minimum values, taking into consideration algorithm effectiveness and practicality.
}
\item{During the content adaptive partition process, the error threshold ${\varepsilon _0}$ decides whether further subdivision of the octree is necessary. In the test, we set the error threshold ${\varepsilon _0}$ to 0.01.
}
\item {
In representing RBF feature functions, we use a Gaussian kernel function for the basis function kernel $\phi$, defined as $\phi  = {e^{ - 0.5{x^2}}}$.}
\item{During the distortion calculation, the octree partition scale $L$ is carefully selected to align with the human visual attention mechanism. $L$ is designated as 16 in the paper.
}
\item{The reference point set we select is all the points in the original normalized point cloud: ${{\bf{P}}_R} = {{\bf{\hat{P}}}_O}$.
}
\end{itemize}
\begin{table}[]
\centering
\setlength{\tabcolsep}{4.0pt}
 \caption{The average rank of different PCQA metrics on 5 datasets. The best and second-best are highlighted in \textcolor{red}{red} and \textcolor{blue}{blue}, respectively.}
 \begin{adjustbox}{max width=\textwidth}
\begin{tabular}{ccccc}
\hline
Criteria     & PLCC                        & SROCC                       & KROCC                       & RMSE                        \\ \hline
GraphSIM     & 5.67                        & 6.56                        & 5.89                        & 5.89                        \\
MS\_GraphSIM & 5.44                        & 7.56                        & 7.00                        & 6.00                        \\
PCQM         & 6.22                        & {\color[HTML]{3531FF} 4.56} & {\color[HTML]{3531FF} 4.89} & {\color[HTML]{3531FF} 4.89} \\
${\text{pSSIM}}_\text{g}$  & 8.25                        & 9.63                        & 9.75                        & 9.13                        \\
${\text{pSSIM}}_\text{c}$  & 8.78                        & 8.11                        & 8.11                        & 8.22                        \\
${\text{MSE}}_{p2po}$     & 9.25                        & 8.00                        & 8.38                        & 7.50                        \\
${\text{PSNR}}_{p2po}$    & 12.88                       & 13.13                       & 12.88                       & 12.38                       \\
${\text{MSE}}_{p2pl}$     & 8.38                        & 6.50                        & 6.38                        & 5.75                        \\
${\text{PSNR}}_{p2pl}$    & 12.25                       & 12.50                       & 12.38                       & 12.75                       \\
${\text{MSE}}_\text{Y}$        & 8.33                        & 6.33                        & 6.11                        & 7.22                        \\
${\text{MSE}}_\text{U}$        & 11.67                       & 11.00                       & 11.56                       & 10.44                       \\
${\text{MSE}}_\text{V}$        & 11.22                       & 10.22                       & 10.33                       & 9.22                        \\
${\text{PSNR}}_\text{Y}$       & {\color[HTML]{3531FF} 5.33} & 5.56                        & 5.33                        & 5.56                        \\
${\text{PSNR}}_\text{U}$       & 8.89                        & 10.89                       & 11.22                       & 11.56                       \\
${\text{PSNR}}_\text{V}$       & 8.33                        & 9.89                        & 10.22                       & 10.56                       \\
\bf{RBFIM}     & {\color[HTML]{FF0000} 3.00} & {\color[HTML]{FF0000} 3.33} & {\color[HTML]{FF0000} 3.33} & {\color[HTML]{FF0000} 4.44} \\ \hline
\end{tabular}
\end{adjustbox}
\label{table3}
\end{table}
\subsection{Performance Comparison}
We compare the performance of each PCQA metric \ADD{regarding distortions caused by both G-PCC and V-PCC compression}. As shown in Table \ref{table2}, the evaluation across five various PCQA datasets. To facilitate \ADD{observations}, the best and \ADD{second-best} performing metrics for each dataset are distinctly highlighted in red and blue within the table, respectively. Moreover, to extensively test the robustness of each metric against different types of compression distortions, we combined the V-PCC and G-PCC distortions from all datasets into a unified dataset. This composite dataset is then divided into two subsets: one comprising all the V-PCC distortions (termed the \text{ALL V-PCC} dataset) and another encompassing all the G-PCC distortions (termed the \text{ALL G-PCC} dataset). Subsequently, these datasets are amalgamated into a comprehensive dataset, labeled the \text{ALL} dataset, to provide a holistic view of metric performance across varied compressed distortion. It is important to mention that the MOS value of the WPC dataset has been converted from a percentage scale to a five-point scale in order to align with the rest of the dataset.\par 
On the whole, according to the results in the databases of the ALL V-PCC dataset and ALL G-PCC dataset, the method proposed in this paper has excellent performance in both G-PCC distortion and V-PCC distortion types and is the best among 16 PCQA indicators. On the ALL dataset, the methods proposed in this paper also showed excellent performance, and PLCC and RMSE showed the best performance among 16 evaluation indicators. SROCC and KROCC performed second best, behind PCQM.\par
Overall, the findings from the comprehensive analysis of the ALL V-PCC and ALL G-PCC datasets indicate that the RBFIM proposed in this research displays superior performance when dealing with both G-PCC and V-PCC distortions. RBFIM outperforms 16 other PCQA metrics, showcasing its robustness across different types of compression distortions. In the ALL dataset, the RBFIM continues to excel, with PLCC and RMSE metrics showing the best performance among the classic and SOTA metrics evaluated. SPLCC and KROCC also performed commendably, ranking second after the PCQM.\par
When evaluating each dataset, the RBFIM always ranks among the top two performing metrics, particularly excelling in G-PCC distortions. However, it is noted that on the M-PCCD dataset, the performance slightly lags behind some multi-feature metrics. This might \ADD{be attributed} to the unique feature of the M-PCCD dataset or the particular distortion type that requires a more nuanced representation of individual features. Metrics that assess multiple features across different scales might capture more details and prove more effective in this case. Further insights reveal that the RBFIM performance on G-PCC distortions is generally more robust compared to V-PCC distortions. This is likely \ADD{because G-PCC leads to more significant alterations in the number and position of points after compression when compared to V-PCC, while the proposed method adapts well to such a situation.} \par
Additionally, a comparative ranking across different datasets is conducted, with each metric's performance averaged and displayed in Table \ref{table3}. The proposed method achieves the highest average rankings across the five datasets in PLCC (3.00), SROCC (3.33), KROCC (3.33), and RMSE (4.44), surpassing both single-feature and multi-feature metrics. PCQM follows closely, securing the second-highest ranks in these metrics. This comparative analysis underlines the \ADD{effectiveness} of the proposed method across various PCQA datasets.\par
The comparative analysis between the performance of different metrics under the ALL dataset is visually represented in Fig. \ref{fig3}. This figure utilizes specific symbols to denote comparative performance: the "\textgreater" symbol (highlighted in green) indicates that the metric on the row outperforms the metric on the column, and the "\textless" symbol (highlighted in yellow) indicates the reverse. In this graphical representation, it is revealed that RBFIM demonstrates superior performance compared to the other 15 metrics in terms of PLCC and RMSE, although it is outperformed by the PCQM metric on SROCC and KROCC.\par
In addition, the findings highlighted in Fig. \ref{fig3} emphasize the effectiveness of multi-feature PCQA metrics over single-feature metrics. This advantage is attributed to the \ADD{capability of} multi-feature metrics to encompass broader and more detailed feature sets, offering a more comprehensive assessment of quality across varying scales.\par
We also compare the average running time of the aforementioned metrics across five point cloud datasets. The tests are conducted on a computer equipped with an Intel(R) Core(TM) i7-8809G CPU @ 3.10G. As shown in \ADD{Table \ref{table4}},\ADD{ RBFIM has less running time compared to all multi-feature methods}. Because this algorithm simplifies the process of establishing correspondences between original point clouds and distorted point clouds by only performing \ADD{calculations of feature function on the distorted point clouds, avoiding the process of searching for corresponding points between the original and distorted point clouds.} This achieves a faster processing speed, particularly suitable for cases with significant geometry quantization. \ADD{Considering} other metrics, due to \ADD{the} simple computational structures, p2po metrics exhibit lower computation complexities\ADD{, while} p2pl metrics have higher complexities due to the need to calculate the normals of each point in the point clouds. GraphSIM and MS-GraphSIM show significantly higher complexities due to key point sampling required in the preprocessing stage. \par
The synthesis of all test results indicates that the proposed RBFIM successfully aligns the distorted point cloud with the original \ADD{in terms of features}. This alignment is important as it \ADD{builds a one-to-one correspondence of features between the two point clouds with different numbers of points}, facilitating a more precise measurement of \ADD{the related} distortion. By improving feature correspondence, \ADD{RBFIM not only aligns but also preserves the intrinsic geometry and topological attributes of point clouds, which are critical for accurate distortion measurement.}\par
\begin{table}[t]
\centering
\setlength{\tabcolsep}{9.5pt}
 \caption{The average running time of each metric on 5 datasets}
\begin{tabular}{cc}
\hline
Criteria     & Running time/s \\ \hline
GraphSIM     & 56.47          \\
MS\_GraphSIM & 72.14          \\
PCQM         & 5.88           \\
${\text{pSSIM}}_\text{g}$  & 16.14          \\
${\text{pSSIM}}_\text{c}$  & 13.31          \\
${\text{MSE}}_{p2po}$     & 5.71           \\
${\text{PSNR}}_{p2po}$    & 5.71           \\
${\text{MSE}}_{p2pl}$     & 12.65          \\
${\text{PSNR}}_{p2pl}$    & 12.65          \\
${\text{MSE}}_\text{Y}$        & 5.71           \\
${\text{MSE}}_\text{U}$        & 5.71           \\
${\text{MSE}}_\text{V}$        & 5.71           \\
${\text{PSNR}}_\text{Y}$       & 5.71           \\
${\text{PSNR}}_\text{U}$       & 5.71           \\
${\text{PSNR}}_\text{V}$       & 5.71           \\
\bf{RBFIM}     & 5.81           \\ \hline
\end{tabular}
\label{table4}
\end{table}
\subsection{The Impact of Key Modules and Parameters}
To verify the reliability of the method, we also test the impact of parameters and key modules involved in the RBFIM.\par
\begin{figure}[]
\centering
\includegraphics[width=3.25in]{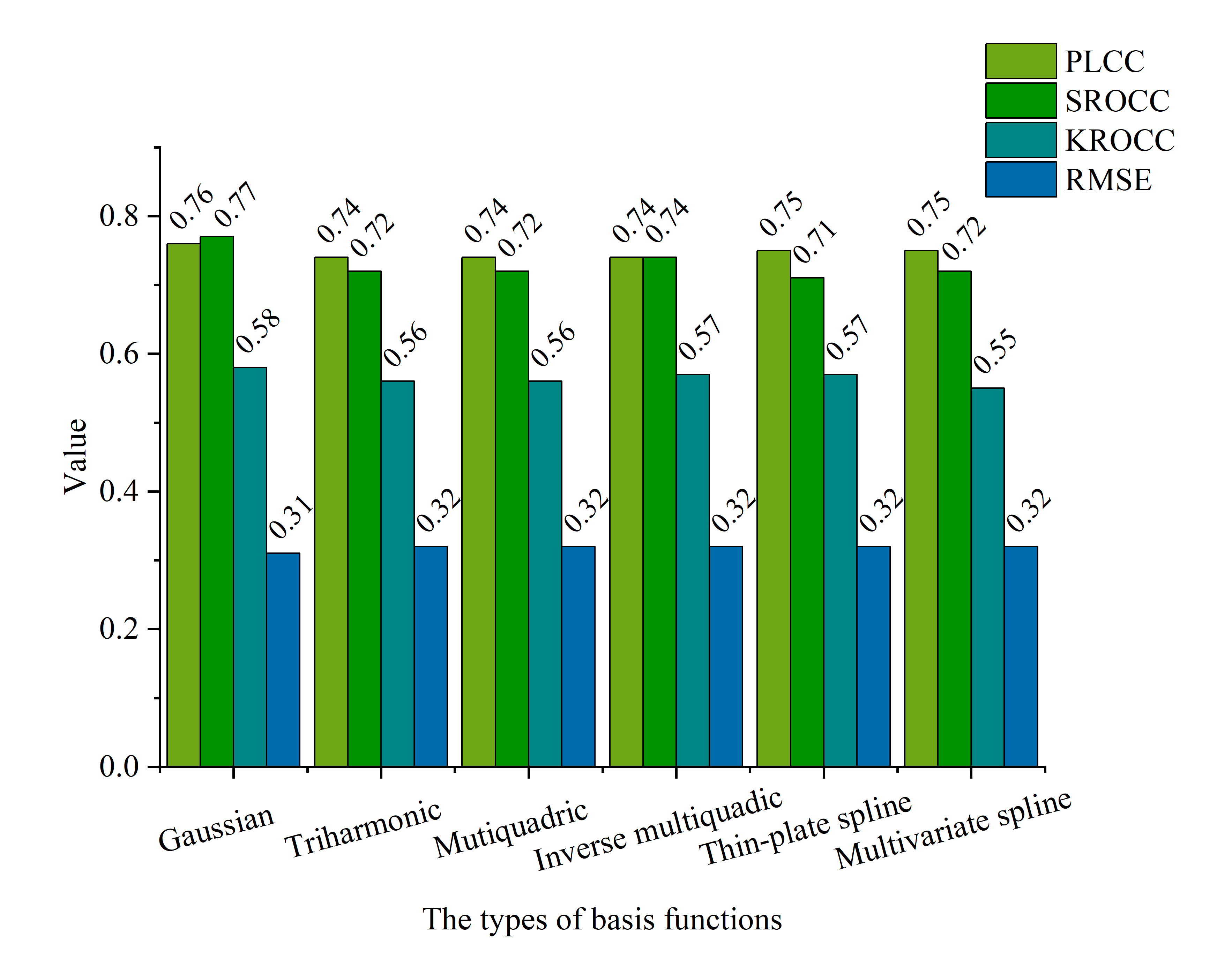}
\caption{\CZADD{Average performance results of different basis function kernels tested on ALL datasets.}}
\label{figure4}
\end{figure}

\begin{itemize}
\item{The influence of RBF kernel}
\end{itemize}
The influence of different types of basis function kernels in RBF is investigated, and the performance of RBFIM is tested on the five datasets using six different types of kernels, as detailed in Table \ref{table5}. The results, illustrated in Fig. \ref{figure4}, indicate that the choice of basis function kernels has little impact on test results. \ADD{The performance is slightly higher using the Gaussian kernel when compared to the other kernel}.\par
\begin{figure}[]
  \centering
  \subfloat[\ADD{PLCC}]{\label{fig:e}\includegraphics[width=2.40in]{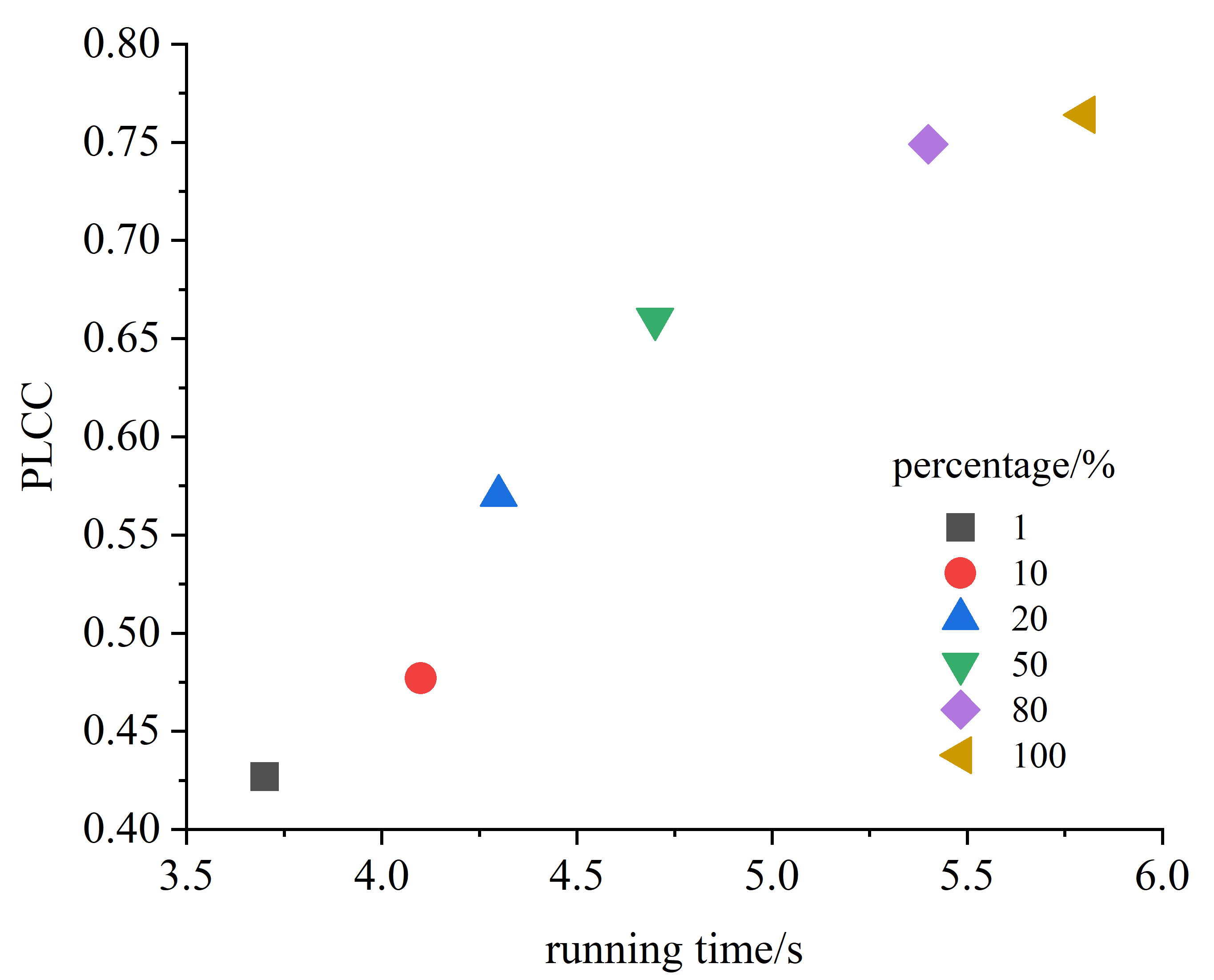}}\quad
  \subfloat[\ADD{Running time}]{\label{fig:i}\includegraphics[width=2.50in]{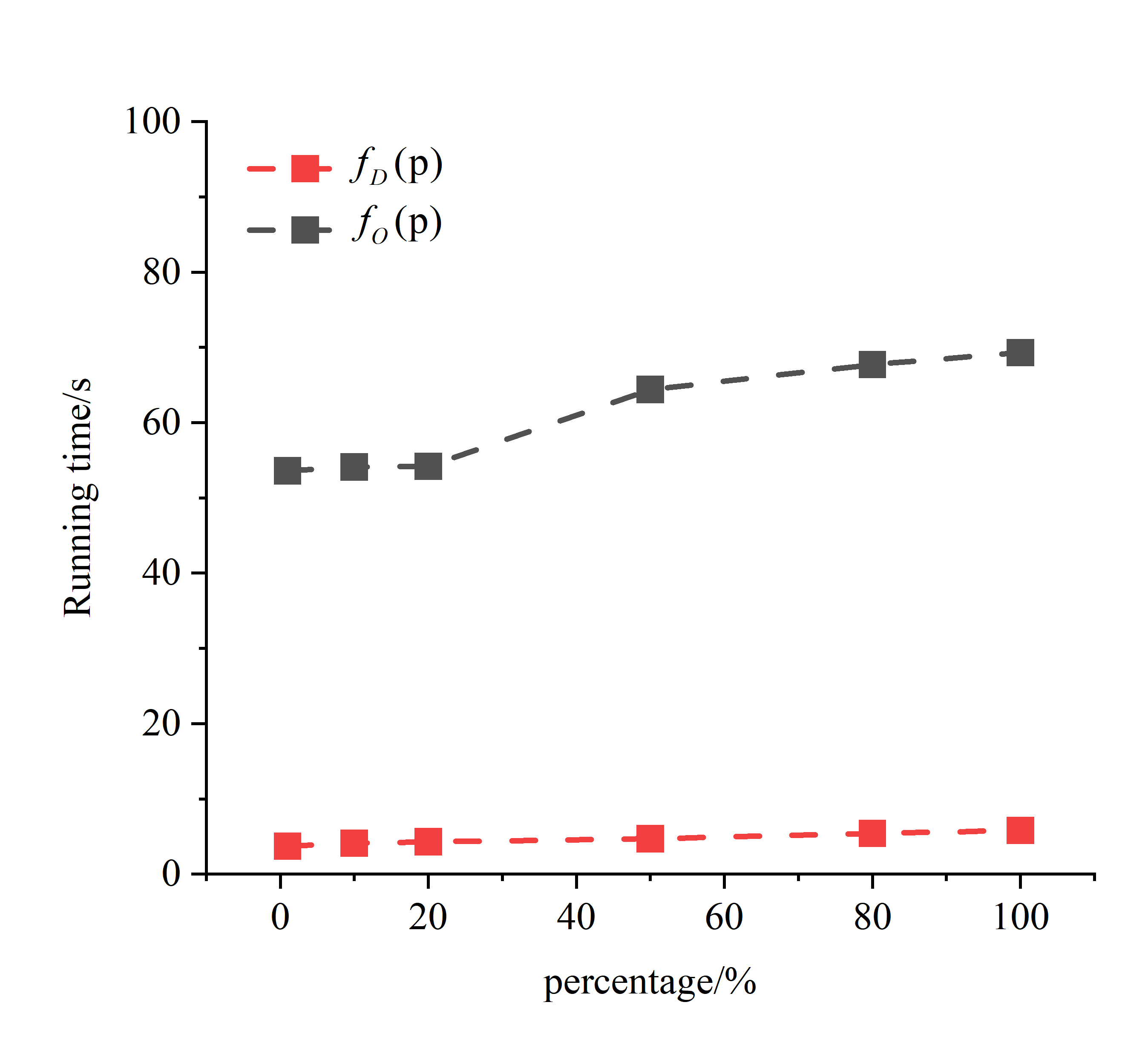}}\\
  \caption{The performance of different reference point sets. \ADD{(a) Comparison of PLCC and running time for using different proportions of the reference point set when the reference point set is the original point cloud. (b) Comparison of runtime for different reference point sets.} ${f_O}({\bf{p}})$ means that the feature function is established by the original point cloud, and the reference point set is from the distorted point cloud. ${f_D}({\bf{p}})$ indicates that the feature function is established by the distorted point cloud, and the reference point set is from the original point cloud.}
  \label{fig5}
  \end{figure}

\begin{table}[]
\centering
\setlength{\tabcolsep}{5.5pt}
 \caption{\ADD{The function forms of different basis function kernels.}}
\begin{tabular}{
>{\columncolor[HTML]{FFFFFF}}c 
>{\columncolor[HTML]{FFFFFF}}c }
\hline
\ADD{The types of kernels} & The function forms \\ \hline
Gaussian                     & $\phi (r) = {e^{ - 0.5{r^2}}}$                  \\
Triharmonic                  & $\phi (r) = {r^3}$                  \\
Mutiquadric                  & $\phi (r) = \sqrt {{r^2} + {{0.5}^2}} $                  \\
Inverse multiquadic          & $\phi (r) = 1/\sqrt {{r^2} + {{0.5}^2}} $                  \\
Thin-plate spline            & $\phi (r) = {r^2}\log (r)$                  \\
Multivariate spline          & $\phi (r) = {r^2}\ln (r)$                  \\ \hline
\end{tabular}
\label{table5}
\end{table}
\begin{itemize}
\item{The impact of \ADD{different} reference point sets}
\end{itemize}
Since the number of points in the reference point set affects the test performance, we evaluate the results at different percentages of the reference point set. \ADD{The test results, as shown in Fig. \ref{fig5} (a), indicate that using different proportions of the reference point set significantly affects the RBFIM performance.} Additionally, we test the performance when establishing feature functions for the original point cloud, using points from the distorted point cloud as reference points. Since the number of points in original point clouds is much higher than that in distorted point clouds, the complexity of establishing feature functions is also higher, \ADD{as shown in Fig. \ref{fig5} (b)}. \par
\begin{table}[]
\centering
\setlength{\tabcolsep}{3.5pt}
 \caption{The performance and time complexity of different feature types on ALL datasets.}
\begin{tabular}{cccccc}
\hline
Feature types & PLCC  & SROCC & KROCC & RMSE  & Running time/s \\ \hline
Luminance     & 0.764 & 0.765 & 0.575 & 0.309 & 5.81           \\
Chroma b      & 0.604 & 0.585 & 0.479 & 0.479 & 5.81           \\
Chroma c      & 0.604 & 0.595 & 0.481 & 0.469 & 5.81           \\
Curvature     & 0.774 & 0.745 & 0.527 & 0.389 & 62.33          \\ \hline
\end{tabular}
\label{table6}
\end{table}
\begin{itemize}
\item{The impact of different feature types}
\end{itemize}
We also test the performance of RBFIM under the selection of different features, as shown in Table \ref{table6}. Under various features, the method proposed exhibits good performance. Since the human eye is more sensitive to \ADD{luma compared to chroma, using luminance as the feature shows better performance. \CZADD{And the results shown in Table \ref{table6} also demonstrated the effectiveness of using curvature as the feature. It is, therefore, concluded that the proposed method works well when the point cloud lacks color information.} However, curvature, which requires additional calculations, adds complexity while achieving outstanding performance}. Moreover, another weakness of \ADD{using} geometry features is that they \ADD{are not useful in} the case when only attributes are lossy. Therefore, this work chooses luminance as the feature. A proper combination of features or using multiple features may further improve the performance of the proposed method, which is currently under investigation. \par

\begin{table}[]
\centering
\setlength{\tabcolsep}{3.5pt}
\caption{\ADD{The performance and time complexity of error thresholds on ALL datasets.}}
\begin{tabular}{cccccc}
\hline
{\color[HTML]{000000} the error thresholds} & {\color[HTML]{000000} PLCC}  & {\color[HTML]{000000} SROCC} & {\color[HTML]{000000} KROCC} & {\color[HTML]{000000} RMSE}  & {\color[HTML]{000000} Running time/s} \\ \hline
{\color[HTML]{000000} 0.1}                  & {\color[HTML]{000000} 0.655} & {\color[HTML]{000000} 0.664} & {\color[HTML]{000000} 0.501} & {\color[HTML]{000000} 0.438} & {\color[HTML]{000000} 7.63}           \\
{\color[HTML]{000000} 0.05}                 & {\color[HTML]{000000} 0.762} & {\color[HTML]{000000} 0.763} & {\color[HTML]{000000} 0.574} & {\color[HTML]{000000} 0.313} & {\color[HTML]{000000} 6.23}           \\
{\color[HTML]{000000} 0.01}                 & {\color[HTML]{000000} 0.764} & {\color[HTML]{000000} 0.765} & {\color[HTML]{000000} 0.575} & {\color[HTML]{000000} 0.309} & {\color[HTML]{000000} 5.81}           \\
{\color[HTML]{000000} 0.001}                & {\color[HTML]{000000} 0.764} & {\color[HTML]{000000} 0.766} & {\color[HTML]{000000} 0.577} & {\color[HTML]{000000} 0.309} & {\color[HTML]{000000} 6.33}           \\
{\color[HTML]{000000} 0.0001}               & {\color[HTML]{000000} 0.763} & {\color[HTML]{000000} 0.763} & {\color[HTML]{000000} 0.574} & {\color[HTML]{000000} 0.314} & {\color[HTML]{000000} 9.93}           \\ \hline
\end{tabular}
\label{table7}
\end{table}
\begin{itemize}
\item{\ADD{The influence of different error thresholds}}
\end{itemize}
\ADD{The error threshold is used to assess the smoothness of the planes formed by the points within the octree, determining whether the octree should continue subdividing. As shown in Table \ref{table7}, when the error threshold gets larger, the number of points in the local interpolation function increases, leading to a decrease in the accuracy of the proposed method. Meanwhile, the increased number of points leads to a higher complexity in solving the linear equations. Conversely, a smaller error threshold results in more frequent octree subdivisions and merges, which also increases the overall computational complexity. Therefore, trading-off between both performance and complexity, we set the error threshold to 0.01.}
\section{Conclusion}\label{sec7}
This paper presents a novel point cloud quality assessment method\ADD{ named RBFIM}. \ADD{By building the feature correspondence using the RBF interpolation function, RBFIM addresses the issue of inaccurate feature correspondence between distorted and original point clouds.} The performance of the proposed algorithm is evaluated on multiple datasets, demonstrating \ADD{advanced performance in both quality assessment and computational complexity} compared to state-of-the-art PCQA metrics. Additional experimental results confirm the model's generalization ability and robustness. \ADD{The proposed method is useful for the optimization of point cloud compression, as well as other processing.} We will also consider more features for RBFIM in future work to establish a more widely used metric.\par

\begin{IEEEbiography}
[{\includegraphics[width=1in,height=1.2in,clip,keepaspectratio]{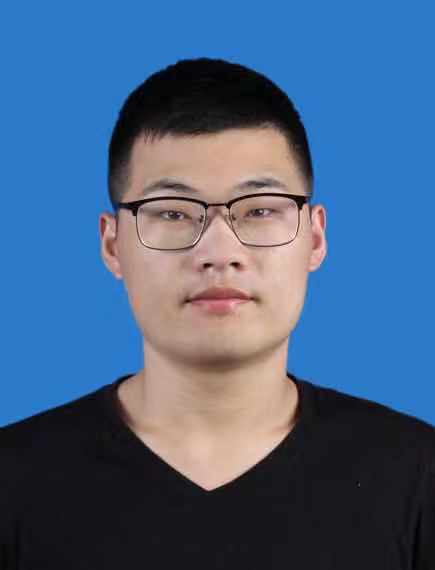}}]{Zhang Chen} received the B.E. degree in 
Electronics and Information Engineering and the M.E. degree in Signal and Information Processing from Northwestern Polytechnical University, Xi’an, China, in 2019 and 2022, respectively. He is currently pursuing the Ph.D. degree in information and Communication Engineering at the same university. His current research interests include point cloud quality assessment, point cloud compression, and 3-D reconstruction.

\end{IEEEbiography}
\begin{IEEEbiography}
[{\includegraphics[width=1in,height=1.2in,clip,keepaspectratio]{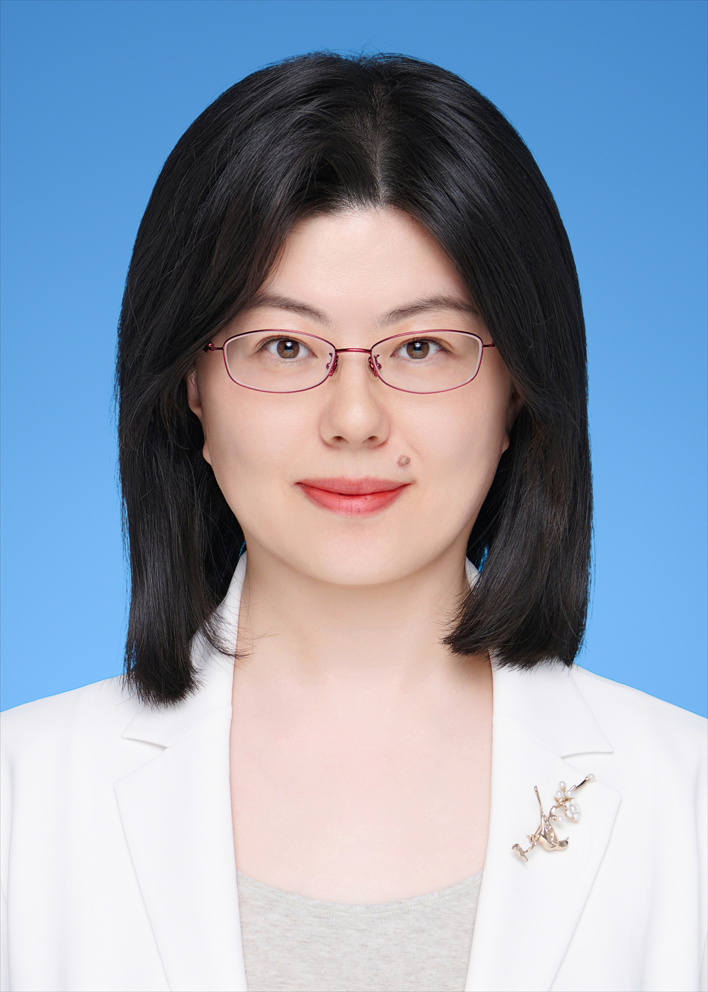}}]{Shuai Wan} (Member, IEEE) received the B.E. degree in Telecommunication Engineering and the M.E. degree in Communication and Information System from Xidian University, Xi’an, China, in 2001 and 2004, respectively, and obtained the Ph.D. in Electronic Engineering from Queen Mary, University of London in 2007. She is now a Professor in Northwestern Polytechnical University, Xi’an, China. Her research interests include scalable/multiview video coding, video quality assessment and hyperspectral image compression.
\end{IEEEbiography}
\begin{IEEEbiography}
[{\includegraphics[width=1in,height=1.2in,clip,keepaspectratio]{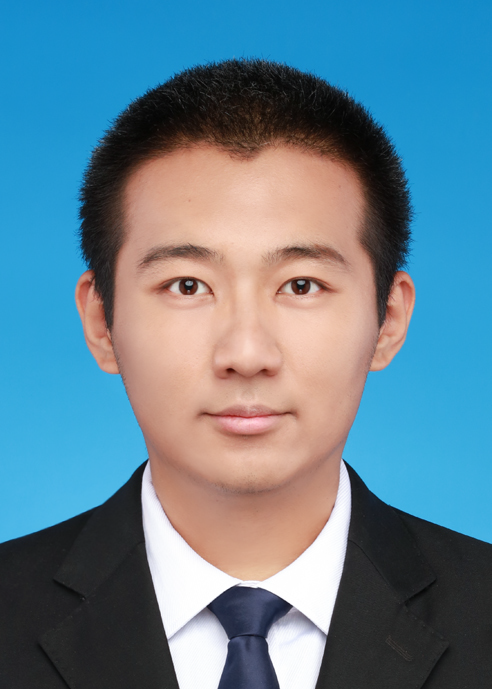}}]{Siyu Ren} received the B.S. degree in Optoelectronic Information Science and Engineering from Tianjin University, Tianjin, China, in 2018. He is currently pursuing the Ph.D. degree in Computer Science at the City University of Hong Kong and Optical Engineering at the Tianjin University. His research interests include deep learning and 3D point cloud processing.
\end{IEEEbiography}
\begin{IEEEbiography}
[{\includegraphics[width=1in,height=1.2in,clip,keepaspectratio]{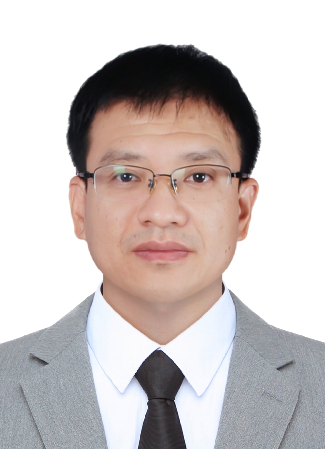}}]{Fuzheng Yang}(Member, IEEE) received the B.E. degree in Telecommunication Engineering, the M.E. degree and the Ph.D. in Communication and Information System from Xidian University, Xi’an, China, in 2000, 2003 and 2005, respectively. He became a lecturer and an Associate Professor in Xidian University in 2005 and 2006, respectively. He has been a professor of communications engineering with Xidian University since 2012. He is also an Adjunct Professor of School of Engineering in RMIT University. During 2006-2007, he served as a visiting scholar and postdoctoral researcher in Department of Electronic Engineering in Queen Mary, University of London. His research interests include video quality assessment, video coding and multimedia communication.
\end{IEEEbiography}
\begin{IEEEbiography}
[{\includegraphics[width=1in,height=1.2in,clip,keepaspectratio]{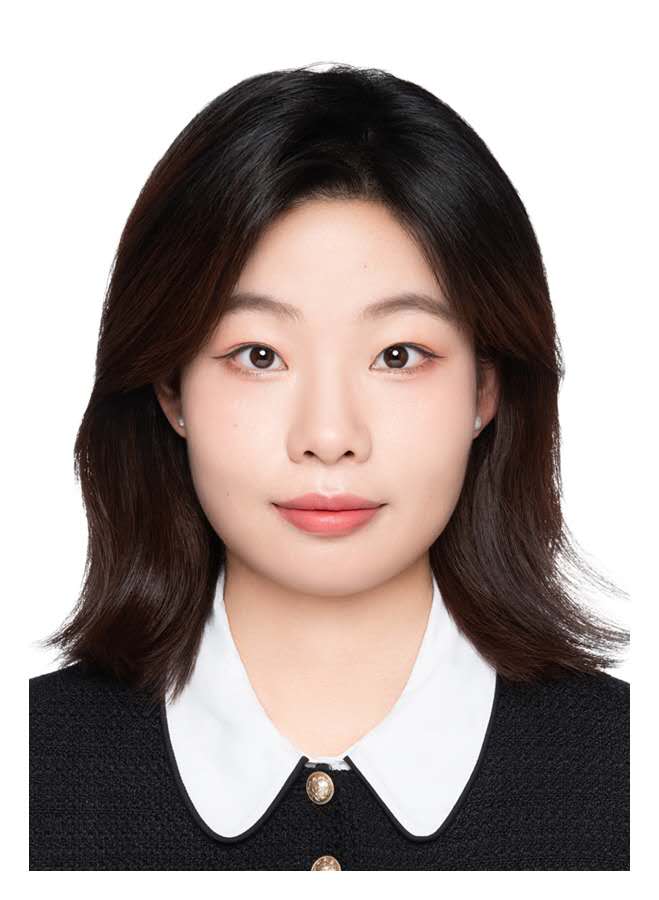}}]{Mengting Yu} is an undergraduate student at the School of Electronics and Information, Northwestern Polytechnical University, Xi'an, China. She will pursue her master's degree at the same institution. Her current research interests focus on point cloud quality assessment, deep learning and point cloud compression.
\end{IEEEbiography}
\begin{IEEEbiography}
[{\includegraphics[width=0.95in,height=1.15in,clip,keepaspectratio]{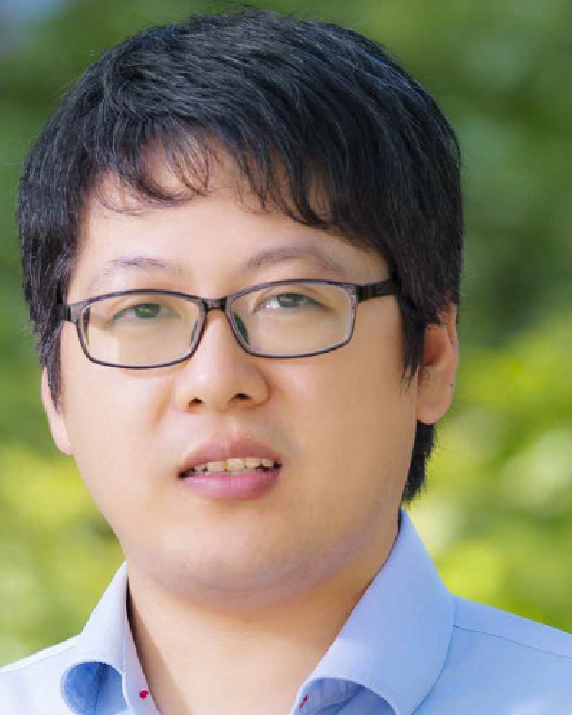}}]{Junhui Hou}(Senior Member, IEEE) is an Asso-
ciate Professor with the Department of Computer
Science, City University of Hong Kong. He holds a
B.Eng. degree in information engineering (Talented
Students Program) from the South China Univer-
sity of Technology, Guangzhou, China (2009), an
M.Eng. degree in signal and information processing
from Northwestern Polytechnical University, Xi’an,
China (2012), and a Ph.D. degree from the School
of Electrical and Electronic Engineering, Nanyang
Technological University, Singapore (2016). His re-
search interests are multi-dimensional visual computing.

Dr. Hou received the Early Career Award (3/381) from the Hong Kong
Research Grants Council in 2018 and the NSFC Excellent Young Scientists Fund in 2024. He has served or is serving as an Associate Editor for IEEE Transactions on Visualization and Computer Graphics, IEEE Transactions on Image Processing, IEEE Transactions on Multimedia, and IEEE Transactions on Circuits and Systems for Video Technology.
\end{IEEEbiography}

\newpage

\end{document}